\documentclass[10pt,twocolumn,letterpaper]{article}

\usepackage{iccv}
\usepackage{times}
\usepackage{epsfig}
\usepackage{graphicx}
\usepackage{amsmath}
\usepackage{amssymb}
\usepackage{multirow}
\usepackage{booktabs}

\usepackage{relsize}
\usepackage{color}
\usepackage[usenames,dvipsnames]{xcolor}
\usepackage{subcaption}
\usepackage[accsupp]{axessibility}  %

\usepackage{pifont}

\usepackage[pagebackref=true,breaklinks=true,letterpaper=true,colorlinks,bookmarks=false]{hyperref}

\iccvfinalcopy %

\def\iccvPaperID{1920} %

\ificcvfinal\fi

\newcommand{\PAR}[1]{\vskip4pt \noindent {\bf #1~}}

\newcommand{\xmark}{\ding{55}}

\newcommand*{\motsynth}{\textit{MOTSynth}\@\xspace}

\begin{document}

\title{MOTSynth: How Can Synthetic Data Help Pedestrian Detection and Tracking? }

\newcommand{\footremember}[2]{%
  \thanks{#2}
    \newcounter{#1}
    \setcounter{#1}{\value{footnote}}%
}
\newcommand{\footrecall}[1]{%
    \footnotemark[\value{#1}]%
}

\author{Matteo Fabbri$^{1,3}$ %
\quad
Guillem Brasó$^2$
\quad
Gianluca Maugeri$^1$
\quad
Orcun Cetintas$^2$
\quad
Riccardo Gasparini$^{1,3}$
\\
\quad
Aljo\u{s}a O\u{s}ep$^2$ 
\quad
Simone Calderara$^1$
\quad
Laura Leal-Taix\'{e}$^2$
\quad
Rita Cucchiara$^1$
\\
$^1$University of Modena and Reggio Emilia, Italy \quad $^2$Technical University of Munich, Germany \quad \\
$^1${\tt\small \{firstname.lastname\}@unimore.it}
\quad $^2${\tt\small \{firstname.lastname\}@tum.de} \\
$^3$GoatAI S.r.l. \\
$^3${\tt\small \{firstname.lastname\}@goatai.it}
\vspace{-10pt}
}

\ificcvfinal\thispagestyle{empty}\fi
\newcommand{\ckm}{\checkmark}
\twocolumn[{%
\renewcommand\twocolumn[1][]{#1}%
\maketitle
\begin{center}
\vspace{-0.5cm}
\includegraphics[width=0.999\linewidth]{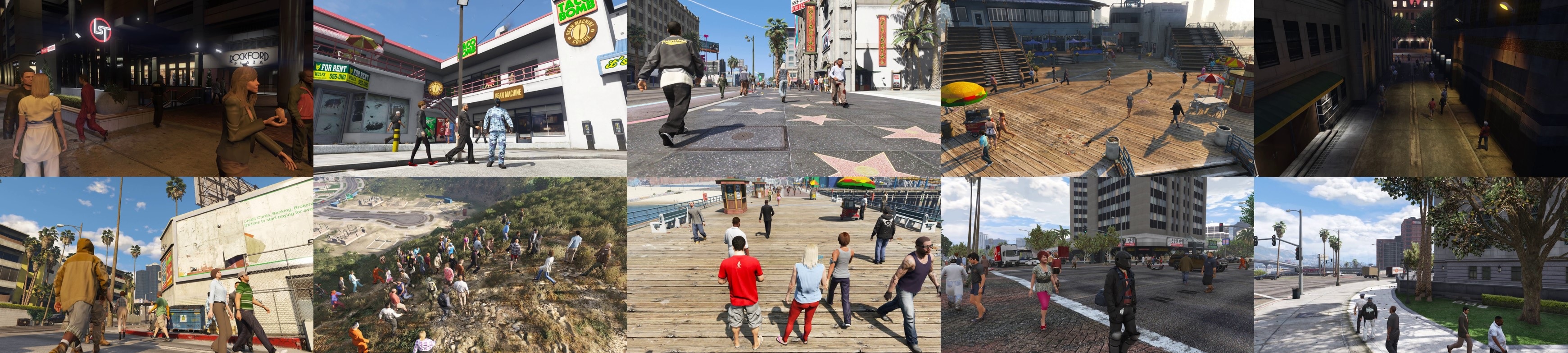}
\vspace{-14pt}
\captionof{figure}{We propose \motsynth, a large and diverse dataset for pedestrian detection, re-identification and multi-object tracking. Due to high diversity, we are able to obtain state-of-the art performance by training models solely on synthetic data.}
\label{fig:teaser}
\vspace{0.1cm}
\end{center}%
}]

\begin{abstract}
   Deep learning-based methods for video pedestrian detection and tracking require large volumes of training data to achieve good performance. 
   However, data acquisition in crowded public environments raises data privacy concerns -- we are not allowed to simply record and store data without the explicit consent of all participants. 
   Furthermore, the annotation of such data for computer vision applications usually requires a substantial amount of manual effort, especially in the video domain. 
   Labeling instances of pedestrians in highly crowded scenarios can be challenging even for human annotators and may introduce errors in the training data. 
   In this paper, we study how we can advance different aspects of multi-person tracking using solely synthetic data. To this end, we generate \motsynth, a large, highly diverse synthetic dataset for object detection and tracking using a rendering game engine. 
   Our experiments show that \motsynth can be used as a replacement for real data on tasks such as pedestrian detection, re-identification, segmentation, and tracking. 
\end{abstract}

\section{Introduction}
\label{sec:experiments}
\noindent

Object detection and tracking in crowded real-world scenarios are challenging and difficult problems with longstanding research history, with applications ranging from autonomous driving to visual surveillance. Since the advent of deep learning, the community has been investigating how to effectively leverage neural networks~\cite{Kim15ICCV, LealTaixe16CVPRW, Milan16arxiv,Son17CVPR, Chu17ICCV, Schulter17CVPR, Kim18ECCV, Fang18WACV, Voigtlaender19CVPR,Bergmann19ICCV,braso2020learning,xu20cvpr,yin2020unified,huang2020sqe} to advance the field. 
However, all these approaches are data-hungry, and data collection and labeling are notoriously difficult and expensive. 
Moreover, dataset collection in public environments\footnote{Crowded public scenes are especially difficult to record during the COVID-19 pandemic.} raises privacy concerns. In fact, European Union already passed privacy-protecting laws such as General Data Protection Regulations (GDPR~\cite{EUdataregulations2018}) to protect the privacy of its citizens that prohibit the acquisition of personal visual data without authorization; ethical issues regarding privacy are also critical in the US, where datasets for training person re-identification modules such as DukeMTMC~\cite{ristani2016performance} were taken offline due to privacy concerns~\cite{harvey19megapixels}.

A possible solution for the aforementioned issues is to employ virtual worlds. The community has already recognized the potential of synthetic data, successfully used for benchmarking~\cite{krahenbuhl2018free} or to compensate for the lack of training data~\cite{amato2019learning, bousmalis2017unsupervised}.
To the best of our knowledge, so far, synthetic data could fully replace recorded data only for low-level tasks such as optical flow estimation~\cite{dosovitskiy2015flownet}. For higher-level tasks, such as object detection, tracking and segmentation, existing methods usually need mixed synthetic and real data and employ alternate training scheme~\cite{amato2019learning} or domain adaptation~\cite{bousmalis2017unsupervised} and randomization~\cite{tobin2017domain} techniques.

In this paper, we aim to answer a challenging question: \textit{Can we advance state-of-the-art methods in pedestrian detection and tracking using only synthetic data?}  
To this end, we created \motsynth, a large synthetic dataset for pedestrian detection, tracking, and segmentation, designed to replace recorded data. \motsynth comes in a bundle with temporally consistent bounding boxes and instance segmentation labels, pose occlusion information, and depth maps. 
As shown in the field of robot reinforcement learning~\cite{tobin2017domain} and vision~\cite{tremblay18CVPRW}, synthetic datasets should significantly vary in terms of lighting, pose, and textures to ensure that the neural network learns all invariances present in the real world. Based on these insights, we generate a large and diverse dataset that varies in terms of environments, camera viewpoints, object textures, lighting conditions, weather, seasonal changes, and object identities (see Fig.~\ref{fig:teaser}). Our experimental evaluation confirms that diversity plays a pivotal role in bridging the synthetic-to-real gap.

The main focus of our study is on \textit{how} \motsynth can help us to advance pedestrian detection, re-identification, and tracking by studying how different aspects of these tasks can benefit from our data.
To this end, we first train several state-of-the-art models for \textit{pedestrian detection}, \textit{segmentation}, \textit{re-identification}, \textit{frame-to-frame regression} and \textit{association} on synthetic data and evaluate their performance on the real-world pedestrian tracking dataset MOTChallenge~\cite{dendorfer20ijcv}. Our experiments show that models, trained on synthetic data are on-par with state-of-the-art on MOTChallenge MOT17\&MOT20, while extremely crowded MOT20 still require fine-tuning. 
Second, we show that prior synthetic datasets~\cite{fabbri2018learning,kim20cvpr} are not suitable for bridging the synth-to-real gap for the task of pedestrian detection and tracking. 
Moreover, we confirm that the diversity in \motsynth is a key for bridging this gap -- and is far more important than the sheer amount of data. 
In addition to a thorough experimental analysis, \motsynth also opens the door to future research on how different components, such as depth and human pose, can be used to advance multi-object tracking in a well-controlled environment.

To summarize, the main contributions of this paper are the following: (i) we open source the largest synthetic dataset for pedestrian detection and tracking with more than $1.3$ million densely annotated frames and $40$ million pedestrian instances; (ii) we show that such a diverse dataset can be a complete substitute for real-world data for high-level tasks such as pedestrian detection and tracking in several scenarios, as well as re-identification and tracking with segmentation; (iii) we provide a comprehensive analysis on how such synthetic worlds can be used to advance the state-of-the-art in pedestrian tracking and detection.

\section{Related Work}
\label{sec:experiments}
\noindent

Advances in computer vision have been driven by the constant growth of available datasets and benchmarks, such as Pascal VOC~\cite{everingham2015pascal}, ImageNet~\cite{russakovsky2015imagenet}, COCO~\cite{Lin14ECCV}, CityScapes~\cite{Cordts16CVPR} and MOTChallenge~\cite{dendorfer20ijcv}. 

\PAR{Multi-object tracking (MOT).} 
In terms of autonomous driving, the pioneering MOT benchmark is the KITTI benchmark~\cite{Geiger12CVPR} that provides labels for object detection and tracking in the form of bounding boxes and segmentation masks~\cite{Voigtlaender19CVPR}. 
However, sequences were collected in a single city in clear weather conditions from a camera mounted on a car. 
The recently proposed BDD100k~\cite{yu20cvpr} covers over 100K videos with high geographic, environmental, and weather diversity. 
Several recent automotive tracking datasets and benchmarks are LiDAR-centric, providing labels in form of 3D bounding boxes~\cite{caesar2020nuscenes,patil2019h3d,sun20CVPR}. 
The recently proposed TAO dataset~\cite{dave20eccv} provides bounding box labels for over 800 object classes.
    
Visual surveillance centric datasets focus on crowded scenarios where pedestrians are interacting and often occluding each other. 
MOTChallenge~\cite{dendorfer20ijcv} benchmark suite played a pivotal role in benchmarking multi-object tracking methods and providing consistently labeled crowded tracking sequences.
In particular, MOT17~\cite{Milan16arxiv} 
provides challenging sequences of crowded urban scenes, capturing severe occlusions and scale variations. MOTS~\cite{Voigtlaender19CVPR} 
The latest release, MOT20~\cite{dendorfer2020mot20} pushes the limits by providing labeled sequences captured in extremely dense scenarios. In terms of car surveillance, UA-DETRAC~\cite{Wen15arxiv} consists of 100 sequences recorded from a high viewpoint with the goal of vehicle tracking. 

Object tracking is deeply entwined with person re-identification (ReID), as several state-of-the-art tracking methods~\cite{Bergmann19ICCV, braso2020learning} rely on learned ReID features. Since DukeMTMC dataset was taken offline due to privacy concerns~\cite{harvey19megapixels}, the most commonly used ReID datasets are Market1501~\cite{zheng2015scalable} and CUHK03~\cite{li2014deepreid}. With this work, we aim to replace recorded data for training object detection, re-identification, and tracking with synthetic data.

\begin{figure*}[t]
\begin{center}
\vspace{-0.5cm}
     \begin{subfigure}[b]{0.33\textwidth}
         \centering
         \includegraphics[width=\textwidth]{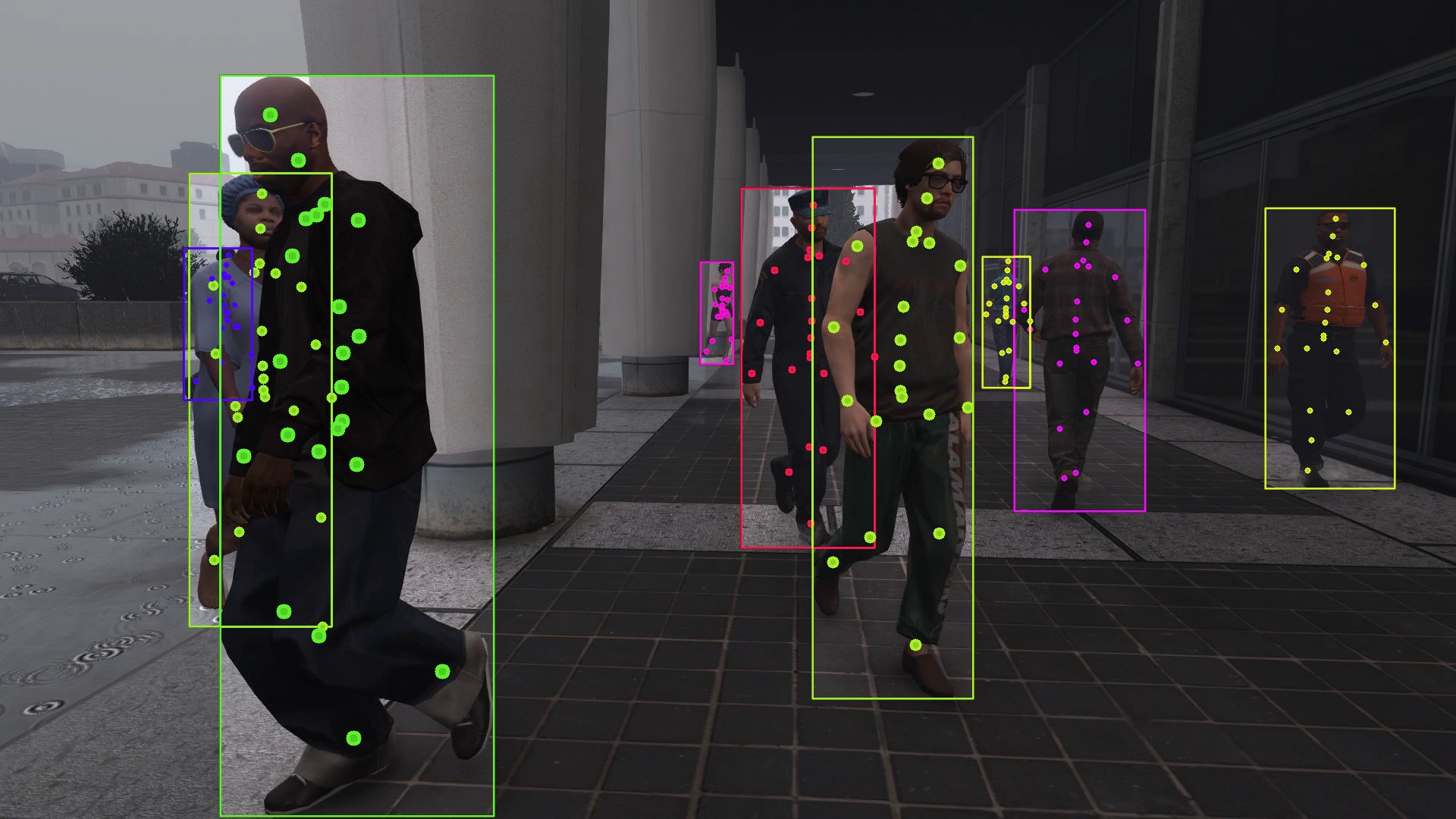}
         \caption{Bounding boxes and pose}
         \label{fig:bbox}
     \end{subfigure}
     \begin{subfigure}[b]{0.33\textwidth}
         \centering
         \includegraphics[width=\textwidth]{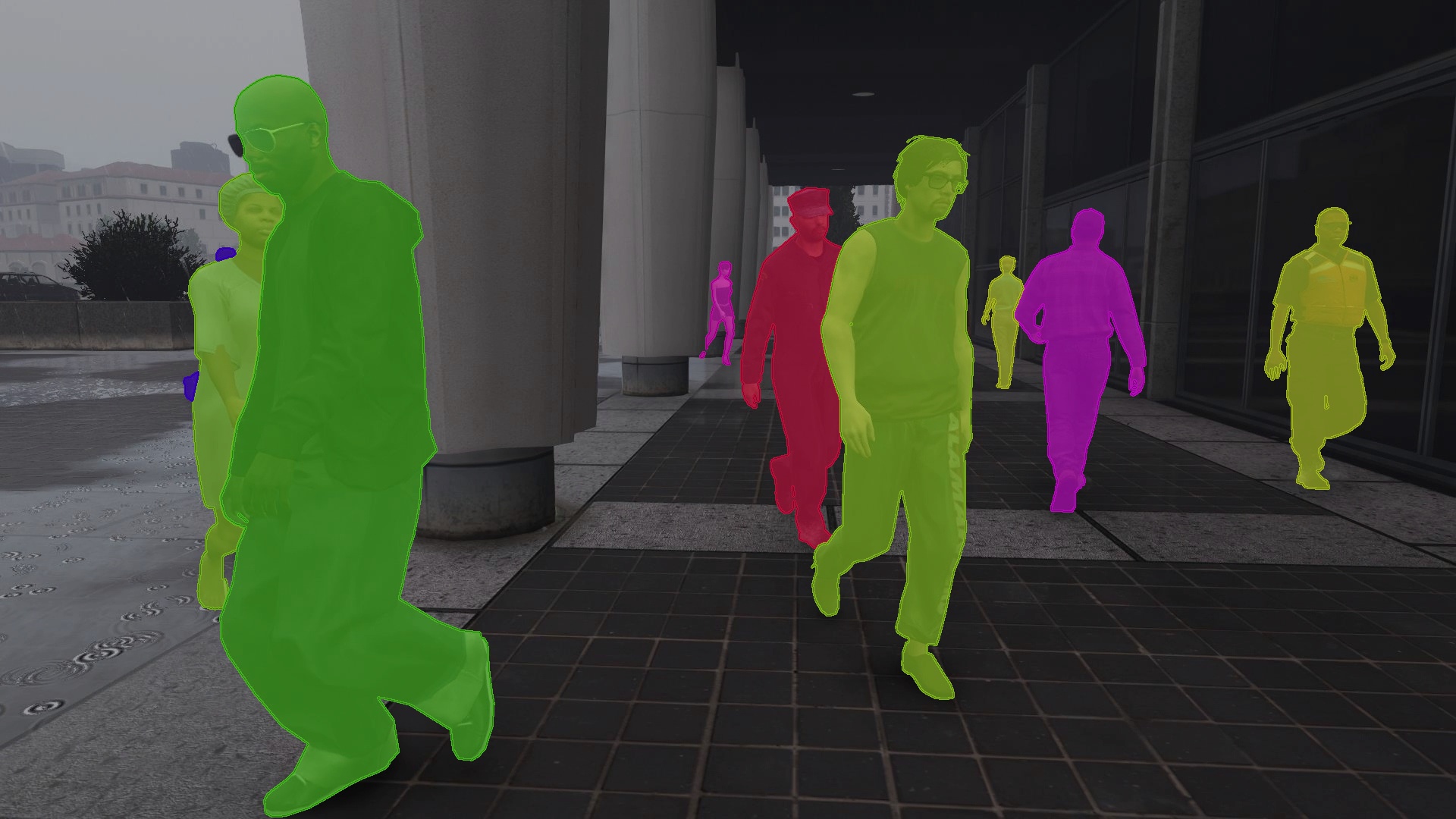}
         \caption{Segmentation masks}
         \label{fig:seg}
     \end{subfigure}
     \begin{subfigure}[b]{0.33\textwidth}
         \centering
         \includegraphics[width=\textwidth]{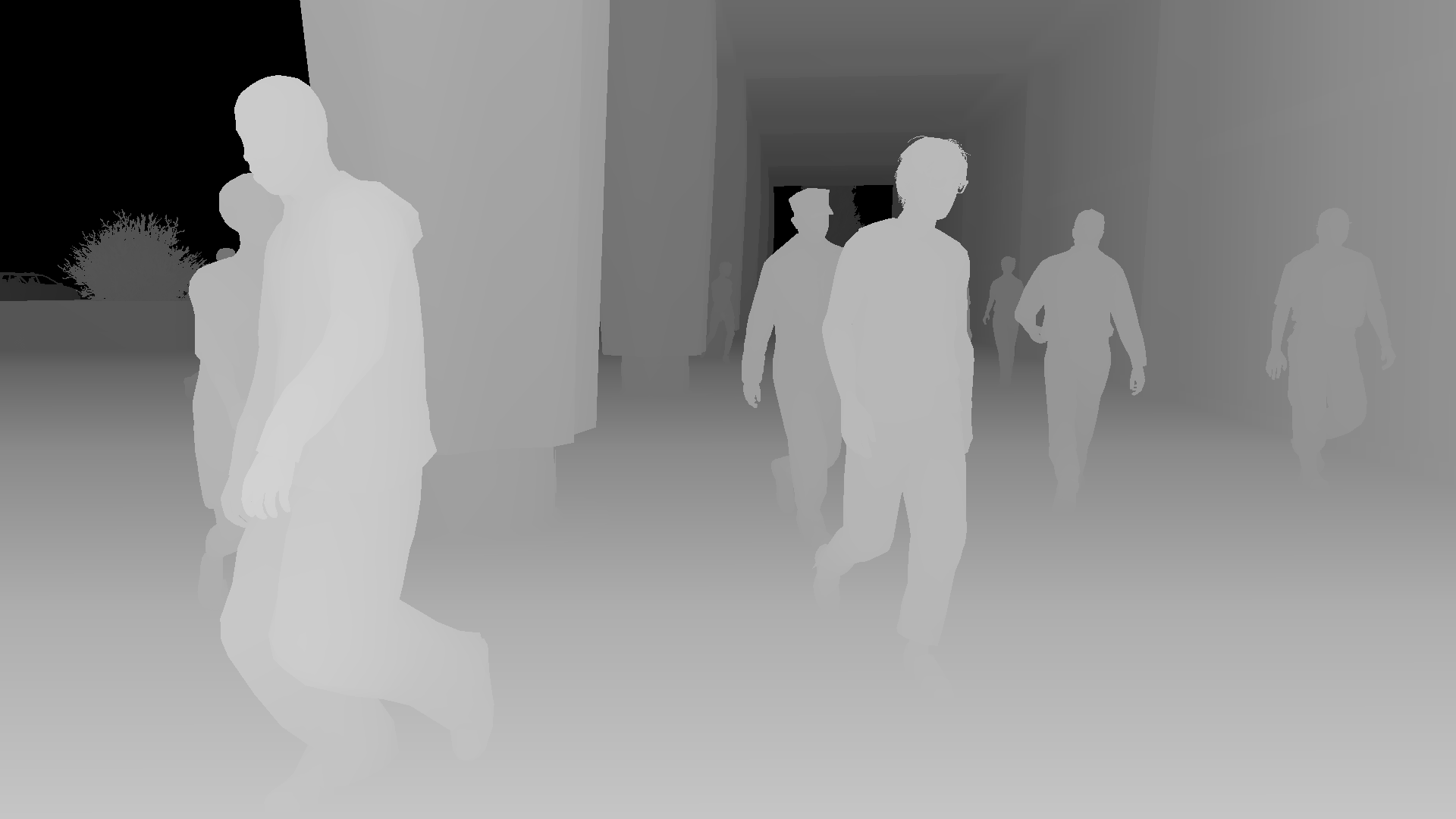}
         \caption{Depth maps}
         \label{fig:depth}
     \end{subfigure}
\vspace{-7pt}
\caption{\motsynth labels. From left to right: bounding boxes and pose, instance segmentation masks, and depth. Best viewed on screen.}
\label{fig:motsynth-labels}
\end{center}
\end{figure*}

\PAR{Synthetic datasets.} 
Data collection usually demands a tremendous amount of manual work. %
As more data is constantly required to train ever-growing models, the cost of labeling such datasets is becoming prohibitive. This burden can either limit the quality or the quantity of available data and hinder progress. A possible solution to the aforementioned problems is to employ virtual worlds. 
Such simulated environments have been successfully applied to low-level tasks, such as feature descriptor computation~\cite{kaneva2011evaluation}, visual odometry~\cite{handa2012real,handa2014benchmark,zhang2016benefit,richter2017playing}, optical flow estimation~\cite{baker2011database,butler2012naturalistic,richter2017playing,mayer2018makes,krahenbuhl2018free,mayer2016large} and depth estimation~\cite{krahenbuhl2018free,mayer2016large}. 
Simulated worlds have also been recently utilized for higher-level tasks like semantic segmentation~\cite{varol2017learning,handa2016understanding,ros2016synthia,hu2019sail,richter2017playing,krahenbuhl2018free,richter2016playing,kim20cvpr}, multi-object tracking~\cite{gaidon2016virtual,fabbri2018learning,sun20eccv, Hu3T19}, hand tracking~\cite{sharp2015accurate}, human pose estimation~\cite{shotton2012efficient,fabbri2018learning,golda2019human,fabbri2020compressed}, 
pedestrian and car detection~\cite{marin2010learning,amato2019learning, johnson2016driving}, and as virtual environments for robot reinforcement learning~\cite{tobin2017domain}. 
The aforementioned works mainly leverage synthetic data for evaluation in scenarios where precise ground-truth data is difficult to obtain~\cite{krahenbuhl2018free} or as means for pre-training data-hungry deep learning models. 
However, apart from optical flow~\cite{dosovitskiy2015flownet}, none of those attempts of using simulated environments was able to replace manually labeled data completely.
In contrast, we focus on bridging the synth-to-real gap for pedestrian detection, ReID, and tracking and perform a thorough analysis of the effect of the amount of training data vs. diversity.

\section{\motsynth Dataset}
\label{sec:dataset}

\motsynth is a large, synthetic dataset specifically designed for training models for pedestrian detection, tracking and segmentation. In the following, we detail the dataset generation process (Sec.~\ref{sec:dataset_generation}) and perform statistical analysis and comparison to other real-world and synthetic datasets (Sec.~\ref{sec:dataset_analysis}).

\subsection{Dataset Generation}
\label{sec:dataset_generation}

To generate \motsynth, we follow prior work~\cite{richter2017playing, kim20cvpr, fabbri2018learning} and we utilize Grand Theft Auto V (GTA-V) video-game, which simulates a city and its inhabitants in a three-dimensional world. 
More precisely, we utilized Script Hook V library~\cite{scripthookv} 

\PAR{Setting up screenplays.}
The first part of recording generation is the scenario (scene) creation. To this end, we manually explored $130 km^2$ (about an eighth of Los Angeles County) of the GTA-V virtual world. 
To generate screenplays, we manually placed camera viewpoints to selected scenarios and set people behavior-related settings, such as the number of pedestrians per scenario, performed actions (such as \textit{standing}, \textit{sitting} or \textit{running}), and paths traveled. In order to simulate dynamics specific to the most crowded areas, we manually pre-planned pedestrian flows by defining a set of trajectories that groups of pedestrians have to follow. We relied on the collision avoidance algorithm to obtain natural pedestrian behavior for each agent. 
For this step, we utilized the mod proposed in \cite{fabbri2018learning} in order to optimize the process. The screenplay generation was the only manual procedure in the creation of \motsynth and took in total only 16 hours. 
To obtain diverse actors, we randomly varied generative attributes of $579$ pedestrian models, provided by the GTA-V game, \eg, different \textit{clothes}, \textit{backpacks}, \textit{bags}, \textit{masks}, \textit{hair} and \textit{beard styles}, yielding over $9,519$ unique pedestrian identities in total. Thus, our generated pedestrians are suitable for training ReID models. We manually set 256 screenplays and combined them with 128 screenplays from~\cite{fabbri2018learning}\footnote{We thank the authors of \cite{fabbri2018learning} for sharing their screenplays.}, totalling 384 screenplays.

\PAR{Rendering.}
After setting up screenplays, we can simulate virtual world dynamics and render different views of the simulated environments. To obtain as diverse renderings as possible, we randomized weather conditions and daytime of the recordings. Weather conditions captured on our dataset are \textit{clear, extra sunny, cloudy, overcast, rainy, thunder, smog, foggy, and blizzard}. 
We recorded each screenplay twice, one during the day and one during the night, totaling 768 generated diverse sequences. 
    
\PAR{Label generation.} Every clip comes with a precise 3D annotation of visible and occluded body parts, temporally consistent 2D bounding boxes and segmentation mask labels for pedestrians, and depth maps (see Fig.~\ref{fig:motsynth-labels}). 
While we do not exploit depth maps in this work, these are cues often used in MOT~\cite{Leibe08TPAMI, Osep17ICRA, Hu3T19, Luiten18ACCV}. Hence, we believe they can be used to further advance the field. 
In terms of completeness, \motsynth exceeds any other dataset in terms of scenario variability, number of entities, and types of annotations. 

\subsection{Statistical Analysis}
\label{sec:dataset_analysis}

\motsynth sequences were rendered as a Full HD video at 25 FPS. 
Each video sequence contains $29.5$ people per frame on average, with a maximum of $125$ people, totaling more than $40M$ bounding boxes over $1.3M$ densely annotated frames. 
The distance of the actors from the camera ranges between $0$ and $101$ meters, resulting in (projected) bounding boxes heights between $0$ and $1,080$ pixels. 

We split \motsynth into training and validation sets, containing $576$ and $192$ clips, respectively. 
We ensured that these splits were roughly balanced in terms of weather conditions, daytime, and density and that no unique person identity appears across these splits.
    
\begin{table}[t]
\begin{center}
\small
\setlength\tabcolsep{2pt}
\begin{tabular}{lrrccccc}
\toprule
\noalign{\smallskip}
Dataset                                     & \#Frames  & \#Inst.   & 3D    & Pose & Segm.   & Depth \\
\noalign{\smallskip}
\hline
\noalign{\smallskip}
PoseTrack~\cite{andriluka2018posetrack} 	& 46k	& 276k 	    &       & \ckm      &               &                     	    \\
MOTS~\cite{Voigtlaender19CVPR}              & 2k	& 26k        &       &           & \ckm          &                	            \\
MOT-17~\cite{Milan16arxiv} 		        & 11k 	& 292k 	    &       & 			&               &                      	    \\
MOT-20~\cite{dendorfer2020mot20} 			& 13k	& 1,652k	&       & 			&               &                      	    \\

\noalign{\smallskip}
\midrule
\noalign{\smallskip}
VIPER~\cite{kim20cvpr}                      & 254k  & 2,750k     & \ckm  &           & \ckm          &               	        \\
GTA~\cite{krahenbuhl2018free}               & 250k   & 3,875k    &       &           & \ckm          & \ckm          	        \\

JTA~\cite{fabbri2018learning}               & 460k	& 15,341k    & \ckm	& \ckm      &               &                      	    \\
\midrule
\midrule
\motsynth 					                & 1,382k	& 40,780k 	& \ckm	& \ckm      & \ckm          &\ckm           	        \\
\bottomrule
\end{tabular}
\vspace{-7pt}
\caption[Publicly available datasets]{Overview of the publicly available datasets for pedestrian detection and tracking. For each dataset, we report the numbers of annotated frames and instances, as well as the availability of different labels.}
\label{table:datasets_small}
\end{center}
\end{table}

In Tab~\ref{table:datasets_small}, we summarize \motsynth statistics in relation to other real and synthetic datasets. In terms of size, the number of instances and labels, \motsynth is superior to all the previously proposed datasets. For a detailed comparison, we refer to the supplementary material.

In contrast to VIPER~\cite{kim20cvpr} and GTA~\cite{krahenbuhl2018free}, \motsynth focuses on crowded pedestrian scenarios. It is larger than pedestrian-focused JTA~\cite{fabbri2018learning} and additionally provides instance segmentation and scene depth information. The key difference between JTA and \motsynth lies in the volume of data, diversity of scenarios, and people variability that, as we experimentally show, allows us to bridge the synth-to-real gap. 

\motsynth contains $40$M bounding boxes with tracking and segmentation mask labels, one to three orders of magnitude more compared to manually labeled MOTChallenge dataset suite (containing $292,733$ bounding boxes in MOT17, $1,652,040$ bounding boxes in MOT20, and $26,894$ segmentation masks in MOTS20 dataset). This difference is most prominent in the case of MOTS20, where pixel-precise labels for pedestrians are hard to obtain, even with semi-automated tools~\cite{Voigtlaender19CVPR}. 

\section{Experimental Evaluation}
\label{sec:experiments}

In this section, we experimentally validate whether \motsynth can be used as a full proxy for (i) pedestrian detection (Sec.~\ref{sec:detection}), (ii) pedestrian re-identification (ReID) (Sec.~\ref{sec:reid}), (iii) multi-object tracking (Sec.~\ref{sec:mot}), and (iv) multi-object tracking and segmentation (Sec.~\ref{sec:mots}). 

\subsection{Experimental Setting}

We evaluate all trained models on the MOTChallenge evaluation suite. To evaluate pedestrian detection and tracking, we use MOT17~\cite{dendorfer20ijcv} and MOT20~\cite{dendorfer2020mot20} datasets. We evaluate our ReID models on MOT17. 
Finally, we evaluate multi-object tracking and segmentation using MOTS20 dataset~\cite{Voigtlaender19CVPR}. 

To understand how well the performance of models trained using synthetic data transfers to the real scenes of MOTChallenge, we train the models using the following datasets for comparison. We make a controlled study using the large-scale COCO dataset~\cite{Lin14ECCV} for detection and tracking, and CrowdHuman~\cite{shao18crowdhuman} for tracking.
For ReID, we employed two real-world ReID datasets Market1501~\cite{zheng2015scalable} and CUHK03~\cite{li2014deepreid}. 

We further compare training on \motsynth with other synthetic datasets depicting humans, namely, JTA~\cite{fabbri2018learning} and VIPER~\cite{richter2017playing}. 
To perform a fine-grained evaluation of \motsynth-to-MOTChallenge transfer capabilities, we split \motsynth into four (inclusive) subsets of $72$, $144$, $288$ and $576$ sequences, named \motsynth-1 to \motsynth-4. This also allows us to study the effect of the amount of data necessary to bridge the synth-to-real gap. For all experiments reported in this paper, we initialize the networks with ImageNet~\cite{Deng09CVPR} pre-trained weights.

\subsection{People Detection}
\label{sec:detection}

\begin{table*}
\center
\tabcolsep=0.11cm
\resizebox{\textwidth}{!}{
\begin{tabular}{c | c | l| c c c| c c c| c c c c c|}
\toprule
& & Dataset & AP $\uparrow$ & MODA $\uparrow$ & FAF $\downarrow$ & TP $\uparrow$ & FP $\downarrow$ & FN $\downarrow$ & Rec. $\uparrow$ & Pr. $\uparrow$  \\ [0.5ex] 
\midrule
\parbox[t]{3mm}{\multirow{20}{*}{\rotatebox[origin=c]{90}{\textbf{MOT17}}}} &
\parbox[t]{3mm}{\multirow{5}{*}{\rotatebox[origin=c]{90}{YOLOv3}}} 
 
& COCO & 69.76 & 62.02 & 1.25 & 47824 & 6650 & 18569 & 72.03 & 87.79\\
\cline{3-11}
& & \motsynth--1 & 62.66 & 52.36 & 1.43 & 42378 & 7613 & 24015 & 63.83 & 84.77 \\
& & \motsynth--2 & 63.08 & 56.67 & 1.22 & 44116 & 6489 & 22277 & 66.45 & 87.18 \\
& & \motsynth--3 & 63.13 & 60.60 & 1.13 & 46264 & 6029 & 20129 & 69.68 & 88.47 \\
& & \motsynth--4 & \textbf{71.90} & \textbf{64.51} & \textbf{1.07} & \textbf{48500} & \textbf{5673} & \textbf{17893} & \textbf{73.05} & \textbf{89.53} \\
\cline{2-11}
& \parbox[t]{3mm}{\multirow{5}{*}{\rotatebox[origin=c]{90}{CenterNet}}} 
& COCO & 67.01 & 44.38 & 3.37 & 47398 & 17935 & 18995 & 71.39 & 72.55\\
\cline{3-11}
& & \motsynth--1 & 61.82 & 49.34 & 2.04 & 43626 & 10866 & 22767 & 65.71 & 80.06 \\
& & \motsynth--2 & 62.32 & 54.90 & \textbf{1.66} & 45269 & \textbf{8820}  & 21124 & 68.18 & \textbf{83.69} \\
& & \motsynth--3 & 62.45 & 55.82 & 1.72 & 46177 & 9117  & 20216 & 69.55 & 83.51 \\
& & \motsynth--4 & \textbf{70.68} & \textbf{57.39} & 1.81 & \textbf{47748} & 9646  & \textbf{18645} & \textbf{71.92} & 83.19 \\
\cline{2-11}
& \parbox[t]{3mm}{\multirow{5}{*}{\rotatebox[origin=c]{90}{Faster R-CNN}}} 
& COCO & 76.68 & 53.86 & \textbf{3.45} & 54127 & \textbf{18364} & 12266 & 81.52 & 74.67\\
\cline{3-11}
& & \motsynth--1 & 76.80 & 39.02 & 5.19 & 53507 & 27603 & 12886 & 80.59 & 65.97 \\
& & \motsynth--2 & 77.47 & 50.62 & 3.82 & 53893 & 20287 & 12500 & 81.17 & 72.65 \\
& & \motsynth--3 & 78.31 & 49.75 & 4.22 & \textbf{55474} & 22441 & \textbf{10919} & \textbf{83.55} & 71.20 \\
& & \motsynth--4 & \textbf{78.98} & \textbf{54.96} & 3.51 & 55121 & 18634 & 11272 & 83.02 & \textbf{74.74} \\
\cline{2-11}
& \parbox[t]{3mm}{\multirow{5}{*}{\rotatebox[origin=c]{90}{Mask R-CNN}}} 
& COCO & 76.96 & 55.55 & 3.31 & 54502 & 17620 & 11891 & 82.09 & 75.57\\
\cline{3-11}
& & \motsynth--1 & 77.58 & 38.43 & 5.51 & 54817 & 29299 & 11576 & 82.56 & 65.17 \\
& & \motsynth--2 & 77.88 & 50.01 & 4.09 & 54930 & 21724 & 11463 & 82.73 & 71.66 \\
& & \motsynth--3 & 78.08 & 49.85 & 4.14 & \textbf{55096} & 21998 & \textbf{11297} & \textbf{82.98} & 71.47 \\
& & \motsynth--4 & \textbf{78.83} & \textbf{56.61} & \textbf{3.17} & 54461 & \textbf{16874} & 11932 & 82.03 & \textbf{76.35} \\
\bottomrule
\end{tabular}
\begin{tabular}{|c | c| l| c c c |c c c| c c }
\toprule
& & Dataset & AP $\uparrow$ & MODA $\uparrow$ & FAF $\downarrow$ & TP $\uparrow$ & FP $\downarrow$ & FN $\downarrow$ & Rec. $\uparrow$ & Pr. $\uparrow$  \\ [0.5ex] 
\midrule
\parbox[t]{3mm}{\multirow{20}{*}{\rotatebox[origin=c]{90}{\textbf{MOT20}}}} &
\parbox[t]{3mm}{\multirow{5}{*}{\rotatebox[origin=c]{90}{YOLOv3}}} 
& COCO & 42.42 & 35.10 & 6.88 & 381635 & 61446 & 530602 & 41.84 & 86.13\\
\cline{3-11}
& & \motsynth--1 & 35.83 & 31.15 & \textbf{2.57} & 307127 & \textbf{22956} & 605110 & 33.67 & 93.05 \\
& & \motsynth--2 & 44.49 & 38.01 & 3.25 & 375739 & 29033 & 536498 & 41.19 & 92.83 \\
& & \motsynth--3 & 44.68 & 42.89 & 3.56 & 423029 & 31797 & 489208 & 46.37 & 93.01 \\
& & \motsynth--4 & \textbf{53.69} & \textbf{48.87} & 2.87 & \textbf{471395} & 25621 & \textbf{440842} & \textbf{51.67} & \textbf{94.85} \\
\cline{2-11}
& \parbox[t]{3mm}{\multirow{5}{*}{\rotatebox[origin=c]{90}{CenterNet}}} 
& COCO & 39.39 & 28.75 & 12.38 & 372835 & 110537 & 539402 & 40.87 & 77.13\\
\cline{3-11}
& & \motsynth--1 & 43.35 & 30.84 & 16.21 & 426095 & 144781 & 486142 & 46.71 & 74.64 \\
& & \motsynth--2 & 43.76 & 40.23 & 7.27 & 431932 & 64971 & 480305 & 47.35 & \textbf{86.92} \\
& & \motsynth--3 & 34.08 & 24.29 & \textbf{6.72} & 281596 & \textbf{60002} & 630641 & 30.87 & 82.43 \\
& & \motsynth--4 & \textbf{51.70} & \textbf{42.18} & 9.72 & \textbf{471592} & 86787 & \textbf{440645} & \textbf{51.70} & 84.46 \\
\cline{2-11}
& \parbox[t]{3mm}{\multirow{5}{*}{\rotatebox[origin=c]{90}{Faster R-CNN}}} 
& COCO & 43.67 & 40.55 & 5.90 & 422649 & 52698 & 489588 & 46.33 & 88.91\\
\cline{3-11}
& & \motsynth--1 & 52.96 & 46.72 & 8.80 & 504790 & 78575 & 407447 & 55.34 & 86.53 \\
& & \motsynth--2 & 52.56 & 46.96 & 7.91 & 498967 & 70609 & 413270 & 54.70 & 87.60 \\
& & \motsynth--3 & 53.37 & 51.38 & 6.36 & 525547 & 56799 & 386690 & 57.61 & 90.25 \\
& & \motsynth--4 & \textbf{53.90} & \textbf{56.03} & \textbf{3.724} & \textbf{544416} & \textbf{33259} & \textbf{367821} & \textbf{59.67} & \textbf{94.25} \\
\cline{2-11}
& \parbox[t]{3mm}{\multirow{5}{*}{\rotatebox[origin=c]{90}{Mask R-CNN}}} 
& COCO & 43.73 & 41.99 & 6.39 & 440081 & 57046 & 472156 & 48.24 & 88.52\\
\cline{3-11}
& & \motsynth--1 & 52.75 & 44.98 & 10.28 & 502154 & 91819 & 410483 & 55.05 & 84.54 \\
& & \motsynth--2 & 53.13 & 50.17 & 6.63 & 516896 & 59225 & 395341 & 56.66 & 89.72 \\
& & \motsynth--3 & 53.51 & 52.27 & 5.76 & 528230 & 51408 & 384007 & 57.90 & 91.13 \\
& & \motsynth--4 & \textbf{54.03} & \textbf{55.69} & \textbf{4.11} & \textbf{544703} & \textbf{36715} & \textbf{367534} & \textbf{59.71} & \textbf{93.69} \\
\bottomrule
\end{tabular}
}
\vspace{-7pt}
\caption{To perform synth-to-real control experiment, we train several object detector models on COCO dataset and on four \motsynth subsets. We evaluate all models on MOTChallenge MOT17 (\textit{left}) and MOT20 (\textit{right}) detection datasets. We observe a clear trend with all object detectors: by purely training on synthetic data, we obtain better performance compared to training on a real-world dataset.}
\label{tab:detection-mot17}
\end{table*}

To understand how training on \motsynth compares to large-scale real-world datasets, we perform a series of experiments involving four heterogeneous object detectors: Faster RCNN~\cite{Ren15NIPS} and Mask RCNN~\cite{He17ICCV} as two-stage detectors, YOLOv3~\cite{redmon2018yolov3} and CenterNet~\cite{zhou2019objects} as single-stage detectors. For each detector, we compare \motsynth training against COCO training by testing on MOTChallenge.

We report the results in terms of average precision (AP), multi-object detection accuracy (MODA~\cite{Bernardin08JIVP}), and the false-positive ratio measured by the number of false alarms per frame (FAF). 
In addition, we report precision, recall, and the absolute number of true positives (TP), false positives (FP), and false negatives (FN). For implementation details on these experiments, we refer to the supplementary. We will focus the discussion on AP as this is the most widely used detection metric. 

\PAR{Synth-to-real transfer.}
As can be seen in Tab.~\ref{tab:detection-mot17}, by training the models on \motsynth, we consistently outperform models trained on COCO. 
When evaluating these models on MOT17, we observe $+2.49$AP improvement for YOLOv3 with \motsynth--4 compared to COCO, $+3.48$AP with CenterNet, $+2.3$AP with Faster R-CNN, and $+1.87$AP with Mask R-CNN. We conclude that the {\it improvements are consistent across different object detectors}. 

These differences are further accentuated on MOT20, where we observe a consistent and remarkable improvement of $+10.97$AP and $+12.31$AP on YOLOv3 and CenterNet, respectively, and $+10.23$AP (Faster R-CNN) and $+10.3$AP (Mask R-CNN). 

We observe that for both MOT17 and MOT20, single-stage detectors benefit from the full \motsynth dataset, while two-stage detectors improve marginally from \motsynth--1 to \motsynth--4 ($+0.12$ and $+0.62$ improvement on MOT17 and $+0.94$ and $+1.28$ improvement on MOT20 in terms of AP with Faster R-CNN and Mask R-CNN, respectively). A possible explanation is that single-stage detectors have to learn a more complex function than two-stage detectors, splitting the problem into two simpler tasks and consequently requiring more data to train effectively. 

\begin{table}
\center
\tabcolsep=0.11cm
\resizebox{\columnwidth}{!}{
\begin{tabular}{c| l| c c c |c c c| c c c c c}
\toprule
& Dataset & AP $\uparrow$ & MODA $\uparrow$ & FAF $\downarrow$ & TP $\uparrow$ & FP $\downarrow$ & FN $\downarrow$ & Rec. $\uparrow$ & Pr. $\uparrow$  \\ [0.5ex] 
 \midrule
\parbox[t]{3mm}{\multirow{4}{*}{\rotatebox[origin=c]{90}{\footnotesize YOLOv3}}} 
& VIPER & 26.65 & 22.02 & \textbf{0.16} & 15447 & \textbf{838} & 50910 & 23.28 & \textbf{94.85}\\
& JTA & 53.18 & 48.77 & 0.79 & 36578 & 4200 & 29815 & 55.09 & 89.70\\
& \motsynth--256 & 62.99 & 62.31 & 0.58 & 44458 & 3090 & 21935 & 66.96 & 93.50 \\
& \textbf{\motsynth-full} & 71.90 & 64.51 & 1.07 & \textbf{48500} & 5673 & \textbf{17893} & \textbf{73.05} & 89.53 \\
\midrule
\parbox[t]{3mm}{\multirow{4}{*}{\rotatebox[origin=c]{90}{\footnotesize CenterNet}}} 
& VIPER & 44.58 & 36.92 & \textbf{1.24} & 31122 & \textbf{6611} & 35271 & 46.88 & \textbf{82.48}\\
& JTA & 60.15 & 45.38 & 2.32 & 42435 & 12308 & 23958 & 63.91 & 77.52\\
& \motsynth--256 & 61.82 & 50.11 & 2.03 & 44067 & 10795 & 22326 & 66.37 & 80.32 \\
& \textbf{\motsynth-full} & \textbf{70.49} & \textbf{55.25} & 2.11 & \textbf{47883} & 11204 & \textbf{18510} & \textbf{72.12} & 81.04 \\
\midrule
\parbox[t]{3mm}{\multirow{4}{*}{\rotatebox[origin=c]{90}{\footnotesize Faster R-CNN}}} 
& VIPER & 60.93 & 42.87 &\textbf{ 2.87} & 43707 & \textbf{15241} & \textbf{10593} & 65.82 & 74.14\\
& JTA & 69.69 & 38.38 & 5.12 & 52726 & 27242 & 13667 & 65.93 & \textbf{79.41}\\
& \motsynth--256 & 78.61 & \textbf{58.65} & 3.10 & \textbf{55441} & 16504 & 10952 & \textbf{83.50} & 77.06 \\
& \textbf{\motsynth-full} & \textbf{78.98} & 54.96 & 3.51 & 55121 & 18634 & 11272 & 83.02 & 74.74 \\
\bottomrule
\end{tabular}}
\vspace{-7pt}
\caption{Comparison on MOT17 against synthetic datasets.}
\label{tab:detection-dataset-comparison}
\end{table}

\PAR{Data volume vs. diversity.} 
To understand the impact of increasing dataset diversity versus increasing the amount of training data, we perform the following experiment. We keep the number of training images fixed and sample images from sequences using two different sampling rates ($1/60$ and $1/10$). The higher the sampling rate, the more images we sample from a given sequence, and vice versa. Thus, by decreasing the sampling rates, we increase the diversity as we sample images from a larger number of different sequences.
When evaluating on the smallest \motsynth--1 subset, we observe a clear trend: {\it diversity matters}. 
When sampling with $1/10$ rate, we reach $76.8$AP and match the COCO model's performance ($76.69$AP). %
However, with denser and therefore less diverse sampling, this is not the case ($70$AP). 
We report detailed results for different object detectors in the supplementary.  

\PAR{Comparison of synthetic datasets.} 
As demonstrated, we were able to bridge the synth-to-real gap using \motsynth. {\it Is this also the case for other synthetic datasets?} To answer this question, we conduct a similar experiment by training models on VIPER~\cite{kim20cvpr} and JTA~\cite{fabbri2018learning} datasets. 
As can be seen in Tab.~\ref{tab:detection-dataset-comparison}, \motsynth-based training clearly outperforms alternative synthetic datasets consistently. 
In particular, YOLOv3 trained on \textbf{\motsynth-full} outperforms VIPER-trained models by $+45.25$AP and JTA trained models by $+18.72$AP. 
We observe a similar trend with CenterNet. We obtain $+25.91$AP improvement with \textbf{\motsynth-full}-trained models over the VIPER model and $+10.34$AP improvement over the JTA model. Finally, with Faster R-CNN, we observe $+18.05$ improvement over the VIPER model and $+18.29$ over the JTA model.

These observations beg the following question: {\it What is the advantage of \motsynth over pedestrian-oriented JTA} -- is it the diversity or sheer amount of data? To answer this question, we conduct the following experiment.  
We train each detector using the subset of \motsynth, \motsynth--256 (\ie, \motsynth--1), containing only 256 sequences, generated from the 128 screenplays provided by the authors of~\cite{fabbri2018learning}. 
The only difference between JTA and \motsynth--256 is in people appearance variation -- high person appearance variety was one of the key goals when generating \motsynth sequences.
As can be seen, with YOLOv3 and Faster R-CNN \motsynth--256 models, we obtain $+9.81AP$ and $+8.92AP$ improvements over JTA trained models. {\it This confirms that the \motsynth diversity in terms of people appearance is a crucial ingredient for bridging the domain gap.}

\begin{table}
\center
\tabcolsep=0.11cm
\resizebox{\columnwidth}{!}{
\begin{tabular}{c| l| c c c |c c c| c c c c c}
\toprule
& Dataset & AP $\uparrow$ & MODA $\uparrow$ & FAF $\downarrow$ & TP $\uparrow$ & FP $\downarrow$ & FN $\downarrow$ & Rec. $\uparrow$ & Pr. $\uparrow$  \\ [0.5ex] 
 \midrule

\parbox[t]{3mm}{\multirow{6}{*}{\rotatebox[origin=c]{90}{MOT17}}} 
& ZIZOM~\cite{lin2018graininess} & \textbf{0.81} & 72.0 & 2.2 & 95414 & 12990 & 19139 & 83.3 & 88.0\\
& SDP~\cite{Yang16CVPR} & \textbf{0.81} & \textbf{76.9} & \textbf{1.3} & 95699 & \textbf{7599} & 18865 & 83.5 & \textbf{92.6}\\
& DPM~\cite{Felzenszwalb08CVPR} & 0.61 & 31.2 & 7.1 & 78007 & 42308 & 36557 & 68.1 & 64.8\\

& FRCNN~\cite{Ren15NIPS} & 0.72 & 68.5 & 1.7 & 88601 & 10081 & 25963 & 77.3 & 89.8\\
\cline{2-10}
& FRCNN \motsynth & 0.80 & 66.7 & 3.7 & 98164 & 21748 & 16400  & 81.9 & 83.7 \\
& FRCNN \motsynth + FT & 0.80 & 71.0 & 3.5 & \textbf{102341} & 20989 & \textbf{12223} & \textbf{89.3} & 83.0 \\
\midrule
\parbox[t]{3mm}{\multirow{4}{*}{\rotatebox[origin=c]{90}{MOT20}}} 
& GNN\_SDT~\cite{wang2020joint} & \textbf{0.81} & \textbf{79.3} & 7.1 & \textbf{304236} & 31677 & \textbf{39288} & \textbf{88.6} & 90.6\\
& VIPeD20~\cite{ciampi2020virtual} & 0.80 & 46.0 & 31.1 & 297101 & 139111 & 46277 & 86.5 & 68.1\\
\cline{2-10}
& FRCNN \motsynth & 0.62 & 52.0 & 6.3 & 206902 & 28202 & 136622 & 60.2 & 88.0 \\
& FRCNN \motsynth + FT & 0.72 & 63.3 & \textbf{5.2} & 241056 & \textbf{23465} & 102468 & 70.2 & \textbf{91.1} \\

\bottomrule
\end{tabular}}
\vspace{-7pt}
\caption{We train Faster R-CNN on \motsynth with and without fine-tuning and evaluate on MOTChallenge MOT17 and MOT20 pedestrian detection test sets.}
\label{tab:detection-bench}
\end{table}

\PAR{Benchmark results.} 
Finally, we evaluate our \motsynth-trained detection models' generalization capability by submitting our results to the MOTChallenge MOT17\&MOT20 benchmarks. We evaluate two variants: no fine-tuning, \ie, trained only on \motsynth, and with fine-tuning (+ FT) on the respective MOTChallenge dataset. We summarize our results in~Tab.~\ref{tab:detection-bench}. 
As can be seen, on MOT17, we outperform (FRCNN \motsynth, $0.8$AP) the baseline Faster R-CNN (FRCNN, $0.72$AP) by $+0.08$AP. 
Interestingly, fine-tuning on the MOTChallenge training set does not significantly impact the \motsynth model in terms of AP. It does, however, improve in terms of MODA ($66.7$ vs. $71$ MODA after fine-tuning), for which a specific threshold needs to be selected. During the experiments, we kept the original threshold. 
It is important to note that more recent object detectors, ZIZOM~\cite{lin2018graininess} and SDP~\cite{Yang16CVPR} only marginally improve over our \motsynth-trained Faster R-CNN models  ($+0.01 AP$). 
    
Unlike MOT17, fine-tuning has a significant effect on MOT20 ($+0.1AP$): we assume this is because in \motsynth we do not have the extremely crowded scenes that are the focus of MOT20. Generating denser synthetic sequences could further help to bridge the gap on MOT20 and remains our future work. 
Finally, we note that detectors specialized in pedestrian detection in crowded scenes~\cite{ciampi2020virtual, wang2020joint} outperform our fine-tuned \motsynth Faster R-CNN model by only $+0.08 AP$.

\subsection{Person Re-Identification}
\label{sec:reid}

To evaluate the re-identification (ReID) model performance, we train three models, (i) trained on Market1501~\cite{zheng2015scalable}, (ii) trained on Market1501~\cite{zheng2015scalable} and CUHK03~\cite{li2014deepreid} and, finally, trained \textit{only} on four subsets \motsynth. 
We evaluate all three models out-of-the-box (without fine-tuning) on the MOTChallenge MOT17 dataset by treating each sequence as a separate dataset. To do so, we randomly select one ground truth box per track to obtain a query set and use the remaining set of boxes, sampled at 10 FPS, as a gallery set. 
We compute standard ReID metrics for every sequence: mean average precision (mAP) and Rank-1 accuracy, and report their average overall sequences. 
All models are trained with a ResNet-50 backbone, followed by a fully connected layer and a standard cross-entropy loss. For implementation details, we refer to the supplementary. 

\begin{table}[t]
\centering
\resizebox{1.0\linewidth}{!}{
\begin{tabular}{c|l|c|c|c}
\toprule
& Dataset  & Split & mAP & Rank1\\
\midrule
\parbox[t]{3mm}{\multirow{2}{*}{\rotatebox[origin=c]{90}{Real}}} 
& Market1501~\cite{zheng2015scalable} & -- & 64.6  & 91.9\\
& Market1501~\cite{zheng2015scalable} + CUHK03~\cite{li2014deepreid} & -- & 69.1 & 91.9\\
\midrule
\parbox[t]{3mm}{\multirow{4}{*}{\rotatebox[origin=c]{90}{Synthetic}}} 
& \multirow{4}{*}{\motsynth}  & 1 & 71.3 & 91.4\\
&  & 2 & 73.1 & 91.8 \\
&  & 3 &  74.2 & 92.6 \\
&  & 4 &  \textbf{75.2} & \textbf{92.8} \\

\bottomrule
\end{tabular}
}
\vspace{-7pt}
\caption{Person ReID experiments on MOT17.} %
\label{tab:reid}
\end{table}

As can be seen in Tab.~\ref{tab:reid}, by training purely on \motsynth data using the first split, we already outperform models trained on real data in terms of mAP ($+6.9$ for Market1501 and $+2.5$ for combined datasets). In terms of Rank1, we obtain $+1.6$ relative to the Market1501-only model. 
However, when training on \motsynth using the first two splits (50\% of total data), we notice an improvement of $+8.6$ and $+4.2$ in terms of mAP and $+3.5$ and $+1$ in terms of Rank1, respectively. {\it This suggests synthetic datasets can be used as a full replacement for ReID datasets, which are often a subject of controversy~\cite{harvey19megapixels}.}

\subsection{Multi-object Tracking }
\label{sec:mot}

In this section, we analyze the value of \motsynth for the task of pedestrian multi-object tracking. 
We report CLEARMOT~\cite{Bernardin08JIVP} and IDF1~\cite{ristani2016performance} metrics and focus the analysis on the most widely used Multiple Object Tracking Accuracy (MOTA) and Identity F1 Score (IDF1). 
We experiment with two different trackers, Tracktor~\cite{Bergmann19ICCV} and recently proposed CenterTrack~\cite{zhou20ECCV}. We evaluate all our models on the most widely used pedestrian tracking dataset, MOTChallenge MOT17~\cite{dendorfer20ijcv}, with and without fine-tuning (\textbf{FT}) on the MOT17 training set. Following the CenterTrack validation scheme, we fine-tune the networks using only the first half of MOT17 sequences and validate on the second half. 

\PAR{Tracktor.} We train detection/tracking~\cite{Bergmann19ICCV} model on (i) COCO dataset~\cite{Lin14ECCV} and (ii) full \motsynth dataset. We note that for Tracktor, we do not need to do any training on sequences, as this method leverages bounding box regression functionality to follow targets. Tracktor also relies on the ReID models to bridge trajectory gaps. 
To this end, we experiment with two ReID models, one trained on real data (Market1501) and one trained on synthetic data (\motsynth).

\PAR{CenterTrack.}
For CenterTrack~\cite{zhou20ECCV} we report the results from the paper. 
In particular, we report (i) CenterTrack model trained on MOT17 directly, the model trained on (ii) the CrowdHuman dataset~\cite{shao18crowdhuman} using a static-image training scheme, and finally, we (iii) report the results we obtain with CenterTrack trained on \motsynth instead of CrowdHuman. 
We train only on the \motsynth--1 subset using full sequences (every frame). 
In this case, we train for four epochs using the same train/eval hyperparameters from the CenterTrack paper. 

\PAR{Fine-tuning.}
We also evaluate how fine-tuning on MOTChallenge affects the final performance of each model. 
In the case of CenterTrack, we employ a slightly modified pre-training scheme to fully utilize the scene diversity of \motsynth. 
Instead of training only on \motsynth--1 subset, we train on all \motsynth sequences. However, due to computational constraints, we only use a subset of frames ($1/8$ of each video) within each sequence. 
This way, we increase the scene diversity while keeping the training time reasonable. 
After training with this subset of \motsynth for 10 epochs, we fine-tune our network on MOT17 sequences for 28 epochs. Throughout fine-tuning and validation, we use the same training and evaluation hyper-parameters as reported in~\cite{zhou20ECCV, Bergmann19ICCV}. 
While \cite{zhou20ECCV} reports performing 70 epochs on the CrowdHuman dataset, we stopped training on \motsynth earlier as we observed our model started over-fitting. For further implementation details of these experiments, we refer to the supplementary.

\begin{table}
\center
\tabcolsep=0.11cm
\resizebox{\columnwidth}{!}{
\begin{tabular}{c|l| c| c |c c c c c c c}
\toprule
& Dataset & FT & ReID & MOTA $\uparrow$ & MOTP $\uparrow$ & IDF1 $\uparrow$ &  TP $\uparrow$ & FP $\downarrow$ & FN $\downarrow$ & IDS $\downarrow$  \\ [0.5ex] 
\midrule
\parbox[t]{3mm}{\multirow{9}{*}{\rotatebox[origin=c]{90}{Tracktor~\cite{Bergmann19ICCV}}}} 
& COCO & \xmark & \xmark & 43.5 & 0.192 & 49.6 & 26783 & \textbf{2816} & 27259 & 467 \\ %
& COCO & \xmark & Market1501 & 44.0 & 0.192 & 55.1 & 26783 & \textbf{2816} & 27259 & 179 \\ %
& COCO & \checkmark & Market1501 & 48.3 & 0.193 & 58.1 & 29218 & 27259 & 24824 & 185 \\ %
\cline{2-11}
& \motsynth & \xmark & \xmark & 45.0 & 0.197 & 51.2 & 28749 & 3992 & 25293 & 458 \\ %
& \motsynth & \xmark & Market1501 & 45.5 & 0.197 & 56.8 & 28749 & 3992 & 25293 & 161 \\
& \motsynth & \xmark & \motsynth & 45.5 & 0.197 & 56.8 & 28749 & 3992 & 25293 & \textbf{160}\\
\cline{2-11}
& \motsynth & \checkmark & \xmark & 49.8 & \textbf{0.199} & 53.8 & \textbf{30588} & 3264 & \textbf{23454} & 411 \\ %
& \motsynth & \checkmark & Market1501 & \textbf{50.3} & \textbf{0.199} & 59.8 & \textbf{30588} & 3264 & \textbf{23454} & 167 \\ %
& \motsynth & \checkmark & \motsynth &\textbf{ 50.3} & \textbf{0.199} & \textbf{59.9} & \textbf{30588} & 3264 & \textbf{23454} & 165 \\ %
\midrule
\midrule
\parbox[t]{3mm}{\multirow{5}{*}{\rotatebox[origin=c]{90}{CenterTr.~\cite{zhou20ECCV}}}} 
& ImageNet & \checkmark & -- & 60.7 & 0.190 & 62.7 & 35443 & 2179 & 18447 & 564 \\
\cline{2-11}
& CrowdHuman & \xmark & -- & 52.2 & \textbf{0.218} & 53.8 & \textbf{32486} & 3604 & 21404 & 728 \\
& CrowdHuman & \checkmark & -- & 66.1 & 0.179 & 64.2 & 38604 & 2442 & 15286 & 528 \\
\cline{2-11}
& \motsynth & \xmark & -- & 54.3 & 0.205 & 57.7 & 33504 & 3601 & 20386 & 666 \\
& \motsynth & \checkmark & -- & \textbf{67.9} & 0.179 & \textbf{66.5} & 38681 & \textbf{1606} & \textbf{15209} & \textbf{508} \\
\bottomrule
\end{tabular}}
\vspace{-7pt}
\caption{Multi-object tracking results performed on MOT17 training set.}
\label{tab:mot}
\end{table}

\PAR{Results.} We report our findings in Tab.~\ref{tab:mot}. 
First, we analyze Tracktor performance. When \textbf{not performing any fine-tuning or using ReID model}, we obtain $45.0$ MOTA and $51.2$ IDF1 with our \motsynth trained model, yielding $+3.5$ MOTA and $+1.6$ IDF1 improvement over the COCO-trained model ($43.5$ MOTA and $49.6$ IDF1).  
After we \textbf{fine-tune} both models on MOT17, the \motsynth-trained model ($49.8$ MOTA, $53.8$ IDF1) improves by $+4.8$ in terms of MOTA and $+2.6$ in terms of IDF1. Similarly, the fine-tuned COCO trained model ($48.3$ MOTA, $58.1$ IDF1) improves by $+4.8$ MOTA and $+8.5$ IDF1. 
After fine-tuning, the improvement of \motsynth over COCO increases by $+1.5$ MOTA and $+4.3$ IDF1, suggesting that {\it \motsynth is more suitable for pre-training pedestrian detection and tracking models compared to COCO dataset.}
   
When \textbf{using ReID models}, we observe consistent improvements on both \motsynth and COCO models. In particular, we observe a consistent improvement in terms of MOTA, which we attribute to a significant reduction ($\sim\!\!250$) in the number of IDS. 
Interestingly, we observe identical improvements with both ReID models, trained on \motsynth and Market1501, and {\it conclude that the ReID model, trained on \motsynth is an adequate replacement for models trained on the real data.} 

CenterTrack reports $60.7$ MOTA and $62.7$ IDF1 when training directly on MOT17, and $52.2$ MOTA and $53.8$ IDF1 on MOT17 when training on CrowdHuman using the static-image training scheme and evaluating directly on MOT17. 
We obtain notably better results when training on \motsynth and evaluating on MOT17: $54.3$ MOTA ($+2.1$) and $57.7$ IDF1 ($+3.9$). 
After fine-tuning the CrowdHuman model on MOT17, we obtain $66.1$ MOTA and $64.2$ IDF1, and even better performance when fine-tuning \motsynth model: $67.9$ MOTA and $66.5$ IDF1. 

Overall, synthetic training always performs favorably, indicating that \motsynth can completely replace manually annotated datasets while increasing performance.

\subsection{Multi-object Tracking and Segmentation}
\label{sec:mots}

In this section, we analyze multi-object tracking and segmentation (MOTS)~\cite{Voigtlaender19CVPR}. We report CLEARMOT metrics~\cite{Bernardin08JIVP}, adapted for MOTS as proposed in~\cite{Voigtlaender19CVPR}. 
Different from MOT, object tracks are localized with segmentation masks instead of bounding boxes. 

\begin{table}
\center
\tabcolsep=0.11cm
\resizebox{\columnwidth}{!}{
\begin{tabular}{c|l| c c c c c c c c}
\toprule

& Dataset & sMOTSA $\uparrow$ & MOTSA $\uparrow$ & MOTSP $\uparrow$ & IDF1 $\uparrow$ & TP $\uparrow$ & FP $\downarrow$ & FN $\downarrow$ & IDS $\downarrow$  \\ [0.5ex] 

\midrule

\parbox[t]{3mm}{\multirow{2}{*}{\rotatebox[origin=c]{90}{\cite{Voigtlaender19CVPR}}}} 
& \multirow{2}{*}{Several} & \multirow{2}{*}{52.74} & \multirow{2}{*}{66.90} & \multirow{2}{*}{80.16} & \multirow{2}{*}{51.19} & \multirow{2}{*}{19202} & \multirow{2}{*}{894} & \multirow{2}{*}{7692} & \multirow{2}{*}{315} \\
&  &  &  &  &  &  &  &  &  \\

\midrule
\parbox[t]{3mm}{\multirow{3}{*}{\rotatebox[origin=c]{90}{\cite{Bergmann19ICCV}}}}
& COCO & 55.58 & 68.80 & \textbf{81.93} & 63.24 & 19677 & 1016 & 7217 & 159 \\
& \motsynth & 55.54 & 68.73 & \textbf{81.93} & 63.09 & 19626 & 987 & 7268 & 155 \\
& \motsynth$^{\dagger}$ & \textbf{56.10} & \textbf{69.52} & 81.67 & \textbf{67.53} & \textbf{19690} & \textbf{850} & \textbf{7204} & \textbf{143} \\
\midrule
\parbox[t]{3mm}{\multirow{2}{*}{\rotatebox[origin=c]{90}{\cite{hornakova20ICML}}}}
& COCO & 53.59 & 67.53 & \textbf{81.52} & 73.31 & 20279 & 2026 & 6615 & 92 \\
& \motsynth & \textbf{54.09} & \textbf{68.07} & 81.49 & \textbf{73.48} & \textbf{20304} & \textbf{1912} & \textbf{6590} & \textbf{86} \\

\midrule
\parbox[t]{3mm}{\multirow{2}{*}{\rotatebox[origin=c]{90}{\cite{zhou20ECCV}}}} 
& COCO & \textbf{53.93} & \textbf{67.49} & 81.68 & \textbf{58.4} & \textbf{19919} & \textbf{1488} & \textbf{6975} & 279 \\
& \motsynth & 53.88 & 67.37 & \textbf{81.74} & 58.19 & 19876 & 1497 & 7018 & \textbf{260} \\

\midrule
\parbox[t]{3mm}{\multirow{2}{*}{\rotatebox[origin=c]{90}{\cite{braso2020learning}}}} 
& COCO & 48.50 & 62.41 & 81.34 & 69.03 & 20061 & 3177 & 6833 & 99 \\
& \motsynth & \textbf{49.17} & \textbf{63.03} & \textbf{81.44} & \textbf{69.40} & \textbf{20079} & \textbf{3047} & \textbf{6815} & \textbf{82} \\

\bottomrule
\end{tabular}}
\vspace{-7pt}
\caption{Multi-object tracking and segmentation. Masks were generated using Mask R-CNN model, trained on COCO and \motsynth. Baselines: Track R-CNN~\cite{Voigtlaender19CVPR}, Tracktor~\cite{Bergmann19ICCV}, Lift\_T~\cite{hornakova20ICML}, CenterNet~\cite{zhou20ECCV}, MPNTrack~\cite{braso2020learning}}
\label{tab:mots}
\end{table}

For these experiments, we first take the tracking outputs of several state-of-the-art MOT methods that use public SDP~\cite{Yang16CVPR} detector and predict segmentation masks with a Mask R-CNN, trained either on COCO or \motsynth. We perform the experiments on the MOTS20 train set and perform no fine-tuning on these sequences. 
In particular, we analyze Tracktor~\cite{Bergmann19ICCV}, Lift\_T~\cite{hornakova20ICML}, CenterNet~\cite{zhou20ECCV}, and MPNTrack~\cite{braso2020learning}. 
For Tracktor, we perform an additional experiment. We turn Mask R-CNN trained on \motsynth into a Tracktor that directly produces pixel-precise tracking output (denoted with $\dagger$). 
For reference, we also report TrackR-CNN~\cite{Voigtlaender19CVPR}, trained on COCO~\cite{Lin14ECCV}, Mapillary Vistas~\cite{Neuhold17ICCV} and MOTS20.

We report our findings in Tab.~\ref{tab:mots}. 
Comparing COCO and \motsynth models, we observe $0.5$ point increase in sMOTSA, MOTSA, and IDF1 for Lif\_T and MPNTrack in favor of \motsynth. 
However, for CenterTrack and Tracktor, we observe a minimal drop in performance ($-0.1$ points). 
Interestingly, Mask R-CNN Tracktor ($\dagger$), trained directly on \motsynth outperforms Tracktor\_v2++ for $+4$ IDF1, $+0.5$ sMOTSA and $+0.7$ MOTSA. This is our best-performing entry on MOTS20. It is important to note that this model is trained \textbf{only} on synthetic data, whereas other methods reported were trained using MOTChallenge, or several datasets in the case of TrackR-CNN. 
We report implementation details for these experiments in the supplementary. 

\subsection{Benchmark Results}
\label{sec:mots}

Finally, we evaluate our models on MOTChallenge MOT17, MOT20, and MOTS20 test sets using the public benchmark. \PAR{MOT17.} As shown in Tab.~\ref{tab:mot-bench}, we obtain highly competitive results when solely training using synthetic data. In fact, on MOT17, Tracktor--\motsynth outperforms Tracktor, trained on COCO, and fine-tuned on MOT17 by $+3.4$ MOTA and $+4.6$ IDF1! Fine-tuning on MOT17 further improves metrics by $+2.2$ MOTA and $+1.9$ IDF1. 
Similarly, CenterTrack trained only on synthetic data achieves competitive results ($59.7$ MOTA, $52.0$ IDF1). The model, fine-tuned on MOT17, further improves performance, establishing a new state of the art with $65.1$ MOTA and $57.9$ IDF1.   

\PAR{MOT20.} On MOT20, Tracktor and CenterTrack trained solely on \motsynth are not yet on-par with state-of-the-art. However, when fine-tuned on MOT20, these models surpass Tracktorv2 by $+3.9$ MOTA and $+0.1$ IDF1. 

\PAR{MOTS20.} We show MOTS20 benchmark results\footnote{In the time of submission, there is only one published entry in MOTS20 benchmark.} in Tab.~\ref{tab:mots-bench}. Our Tracktor$^{\dagger}$ Mask R-CNN trained \textbf{only} on synthetic data significantly outperforms TrackR-CNN~\cite{Voigtlaender19CVPR}, that is trained on COCO, Mapillary Vistas~\cite{Neuhold17ICCV} and MOTS20 training set. In particular, we improve sMOTSA for $+14.45$, MOTSA for $+15.04$, MOTSP for $+3,47$ and IDF1 for $+21.47$. This confirms our intuition that \motsynth is especially beneficial in the scarce data regime, as is the case for the MOTS task. 

These experiments confirm that top-ranked MOTChallenge models can be trained purely on synthetic data on the MOT17 and MOTS datasets to achieve state-of-the-art results. However, on the MOT20 dataset, fine-tuning is still needed to reach state-of-the-art results. We assume that this is due to the fact that synthetic sequences more closely resemble MOT17 sequences than extremely crowded MOT20 sequences. This hints that \motsynth has future potential in closing this gap by simulating similarly dense environments.

\begin{table}
\center
\tabcolsep=0.11cm
\resizebox{\columnwidth}{!}{
\begin{tabular}{c|l| c c c | c c c}
\toprule
& Method & MOTA $\uparrow$ & MOTP $\uparrow$ & IDF1 $\uparrow$ & FP $\downarrow$ & FN $\downarrow$ & IDS $\downarrow$  \\ [0.5ex] 
\midrule
\parbox[t]{3mm}{\multirow{10}{*}{\rotatebox[origin=c]{90}{MOT17}}} 
& Tracktor-\motsynth & 56.9 & 78.0 & 56.9 & 20852 & 220273 & 2012 \\
& Tracktor-\motsynth + FT & 59.1 & 79.3 & 58.8 & 22231 & 206062 & 2323 \\
& Tracktor~\cite{Bergmann19ICCV} & 53.5 & 78.0 & 52.3 & 12201 & 248047 & 2072  \\
& Tracktorv2~\cite{Bergmann19ICCV} & 56.3 & 78.8 & 55.1 & \textbf{8866} & 235449 & 1987 \\
\cline{2-8}
& CenterTrack-\motsynth & 59.7 & 77.4 & 52.0 & 39707 & 181471 & 6035 \\
& CenterTrack-\motsynth + FT & \textbf{65.1} & \textbf{79.9} & 57.9 & 11521 & \textbf{180901} & 4377 \\
& CenterTrack~\cite{zhou20ECCV} & 61.5 & 78.9 & 59.6 & 14076 & 200672 & 2583 \\
\cline{2-8}

& Lif\_T~\cite{hornakova20ICML} & 60.5 & 78.3 & 65.6 & 14966 & 206619 & 1189  \\
& LPC~\cite{dai2021learning} & 59.0 & 78.0 &\textbf{ 66.8} & 23102 & 206948 & \textbf{1122}  \\
& MPNTrack~\cite{braso2020learning} & 58.8 & 78.6 & 61.7 & 17413 & 213594 & 1185  \\

\midrule
\midrule
\parbox[t]{3mm}{\multirow{8}{*}{\rotatebox[origin=c]{90}{MOT20}}} 
& Tracktor-\motsynth & 43.7 & 75.1 & 39.7 & 15933 & 271814 & 3467 \\
& Tracktor-\motsynth + FT & 56.5 & 78.8 & 52.8 & 11377 & 211772 & 1995 \\
& Tracktorv2~\cite{Bergmann19ICCV} & 52.6 & 79.9 & 52.7 & \textbf{6930} & 236680 & 1648 \\
\cline{2-8}
& CenterTrack-\motsynth & 39.7 & 72.9 & 37.2 & 47066 & 259274 & 5872 \\
& CenterTrack-\motsynth + FT & 41.9 & \textbf{80.2} & 38.2 & 36594 & 258874 & 5313 \\
\cline{2-8}
& MPNTrack~\cite{braso2020learning} & \textbf{57.6 }& 79.0 & 59.1 & 16953 & \textbf{201384} & \textbf{1210}  \\
& LPC~\cite{dai2021learning} & 56.3 & 79.7 & \textbf{62.5} & 11726 & 213056 & 1562  \\
& SORT20~\cite{bewley2016simple} & 42.7 & 78.5 & 45.1 & 27521 & 264694 & 4470  \\
\bottomrule
\end{tabular}}
\vspace{-7pt}
\caption{Benchmark results on MOT17 and MOT20. We refer to supplementary for the full version.}
\label{tab:mot-bench}
\end{table}

\begin{table}
\center
\tabcolsep=0.11cm
\resizebox{\columnwidth}{!}{
\begin{tabular}{l| c c c c c c c c}
\toprule

Method & sMOTSA $\uparrow$ & MOTSA $\uparrow$ & MOTSP $\uparrow$ & IDF1 $\uparrow$ & TP $\uparrow$ & FP $\downarrow$ & FN $\downarrow$ & IDS $\downarrow$  \\ [0.5ex] 

\midrule

Tracktor$^{\dagger}$-\motsynth & \textbf{55.1} & \textbf{70.2} & \textbf{79.6} & \textbf{63.9} & \textbf{23994} & \textbf{1128} & \textbf{8275} & \textbf{200} \\
TrackRCNN~\cite{Voigtlaender19CVPR} & 40.6 & 55.2 & 76.1 & 42.4 & 19628 & 1261 & 12641 & 567 \\
\bottomrule
\end{tabular}}
\vspace{-7pt}
\caption{Benchmark results on MOTS20 dataset.}
\label{tab:mots-bench}
\end{table}

\section{Conclusion}
\label{sec:experiments}

We presented \motsynth, a massive synthetic dataset for pedestrian detection and tracking in urban scenarios. We experimentally demonstrated that synthetic data could completely substitute real data for high-level in-the-wild scenarios, such as pedestrian detection, re-identification tracking, and segmentation. Remarkably, we obtained state-of-the-art results on the MOTChallenge MOT17 dataset by training recent methods using solely synthetic data. 
We believe this paper will pave the road for future efforts in replacing costly data collection with synth in other domains. 

\PAR{Acknowledgments.}
\small{
The authors would like to thank Tim Meinhardt for his helpful comments on our manuscript. 
For partial funding of this project, GB, AO, OC, and LLT would like to acknowledge the Humboldt Foundation through the Sofja Kovalevskaja Award. MF, GM, RG, SC, and RC would like to acknowledge the InSecTT project, funded by the ECSEL Joint Undertaking (JU) under GA 876038. The JU receives support from the EU H2020 Research and Innovation programme and AU, SWE, SPA, IT, FR, POR, IRE, FIN, SLO, PO, NED, TUR. The document reflects only the author’s view and the Commission is not responsible for any use that may be made of the information it contains.
}

{\small
\bibliographystyle{ieee_fullname}
\bibliography{abbrev_long,refs}
}

\clearpage
\appendix
\part*{Supplementary Material}

\begin{table*}[ht!]
\begin{center}
\small
\setlength\tabcolsep{4pt}
\begin{tabular}{lrrrcccccccc}
\toprule
\noalign{\smallskip}
Dataset                                     & \#Clips   & \#Frames  & \#Instances   & 3D    & Occl. & Pose Est. & Inst. Segm.   & Depth Est.      & Type      \\
\noalign{\smallskip}
\hline
\noalign{\smallskip}
KITTI~\cite{Geiger12CVPR} 		            & 50 		& 22,000 	& 160,000 	    & \ckm  & \ckm  &           &               &  \ckm         	    & AD        \\ 
nuSCENES~\cite{caesar2020nuscenes} 		    & 1,000  	& 40,000 	&   280,000     & \ckm	& 	    &           &               &                         	    & AD        \\
BDD100k-MOTS~\cite{Yuan183DV}               & 70        & 14,000    & 129,000       &       & \ckm	&           & \ckm          &                 	            & AD        \\
BDD100k-MOT~\cite{Yuan183DV}                & 1,600     & 100,000   & 3,300,000     &       & \ckm	&           &               &                 	            & AD        \\
Waymo Open~\cite{sun20CVPR}                 & 1,150     & 230,000	& 2,700,000     & \ckm  &  	    &           &               &                 	            & AD        \\
\noalign{\smallskip}
\midrule
\noalign{\smallskip}
TAO~\cite{dave20eccv}                       & 2,907	    & 148,235	& 175,723       &       &  	    &           &               &                 	            & DV   \\
PoseTrack~\cite{andriluka2018posetrack} 	& 1,356  	& 46,000 	& 276,000 	    &       &       & \ckm      &               &                      	    & DV   \\
MOTS~\cite{Voigtlaender19CVPR}              & 4	        & 2,862	    & 26,894        &       & \ckm	&           & \ckm          &                 	            & US     \\
MOT-17~\cite{milan2016mot16} 		        & 14 		& 11,235 	& 292,733 	    &       & \ckm	& 			&               &                      	    & US     \\
MOT-20~\cite{dendorfer2020mot20} 			& 8 		& 13,410 	& 1,652,040 	&       & \ckm	& 			&               &                      	    & US     \\
\noalign{\smallskip}
\midrule
\noalign{\smallskip}
VIPER~\cite{kim20cvpr}                      & 187	    & 254,064   & 2,750,000     & \ckm  & \ckm	&           & \ckm          &               	        & AD        \\
GTA~\cite{krahenbuhl2018free}              & -	        & 250,000   & 3,875,000     &       &  	    &           & \ckm          & \ckm          	        & DV   \\
JTA~\cite{fabbri2018learning}               & 512 		& 460,800	& 15,341,242    & \ckm	& \ckm 	& \ckm      &               &                      	    & US     \\
\midrule
\midrule
\motsynth 					                & 768 		& 1,382,400	& 40,780,800 	& \ckm	& \ckm 	& \ckm      & \ckm          &\ckm            	        & US     \\
\bottomrule
\end{tabular}
\captionof{table}{Overview of the publicly available datasets for pedestrian detection and tracking. For each dataset, we report the numbers of clips, annotated frames and instances. We also report the presence of 3D data and occlusion information, as well as the availability of labels for pose estimation, instance segmentation, and depth estimation. The last column shows the data type: autonomous driving (AD), diverse (DV) or urban surveillance (US).}
\label{table:datasets}
\vspace{0.1cm}
\end{center}
\end{table*}

\begin{figure*}[ht]
\begin{center}
\vspace{-0.5cm}
     \begin{subfigure}[b]{0.19\textwidth}
         \centering
         \includegraphics[width=\textwidth]{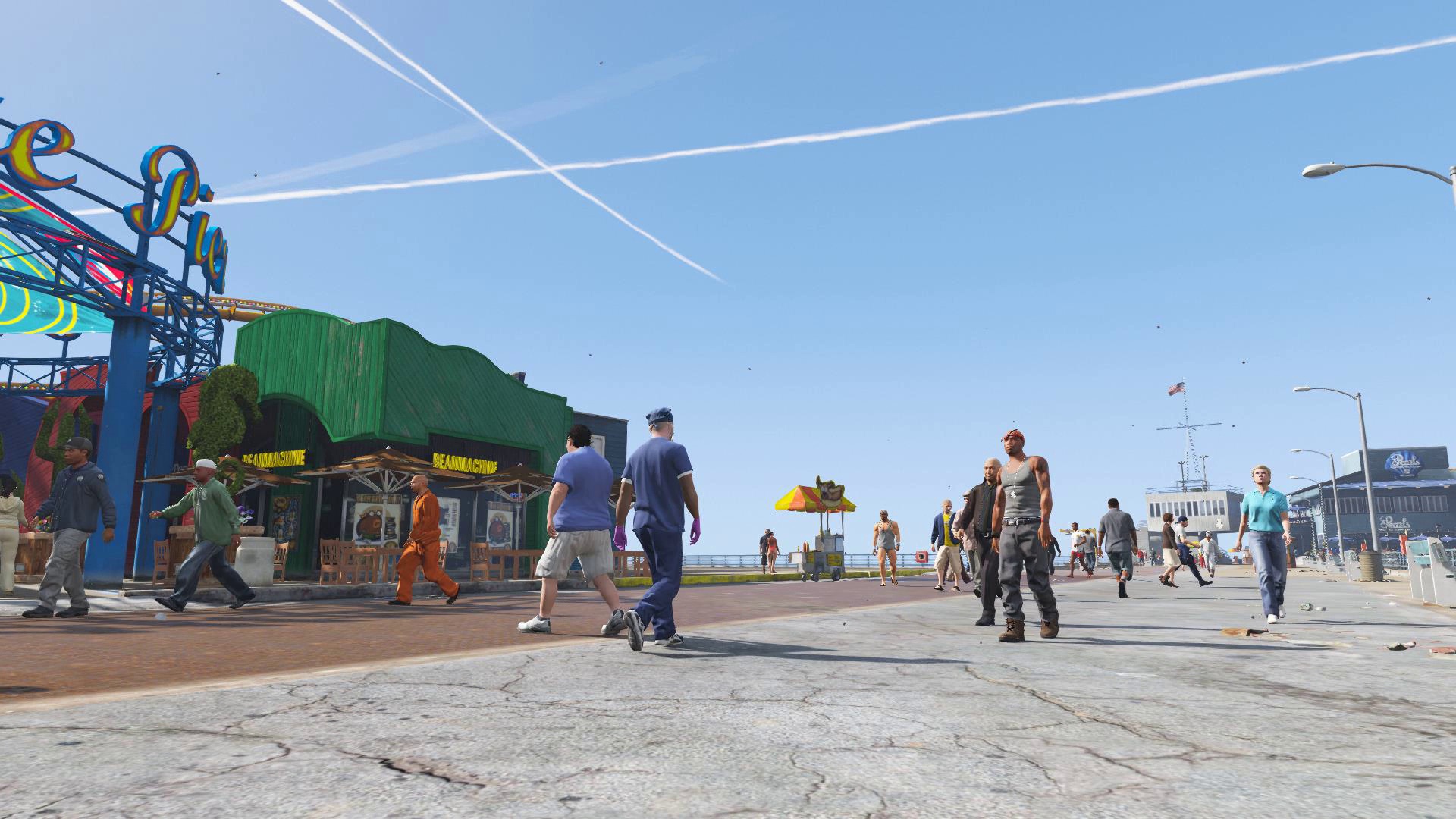}
         \includegraphics[width=\textwidth]{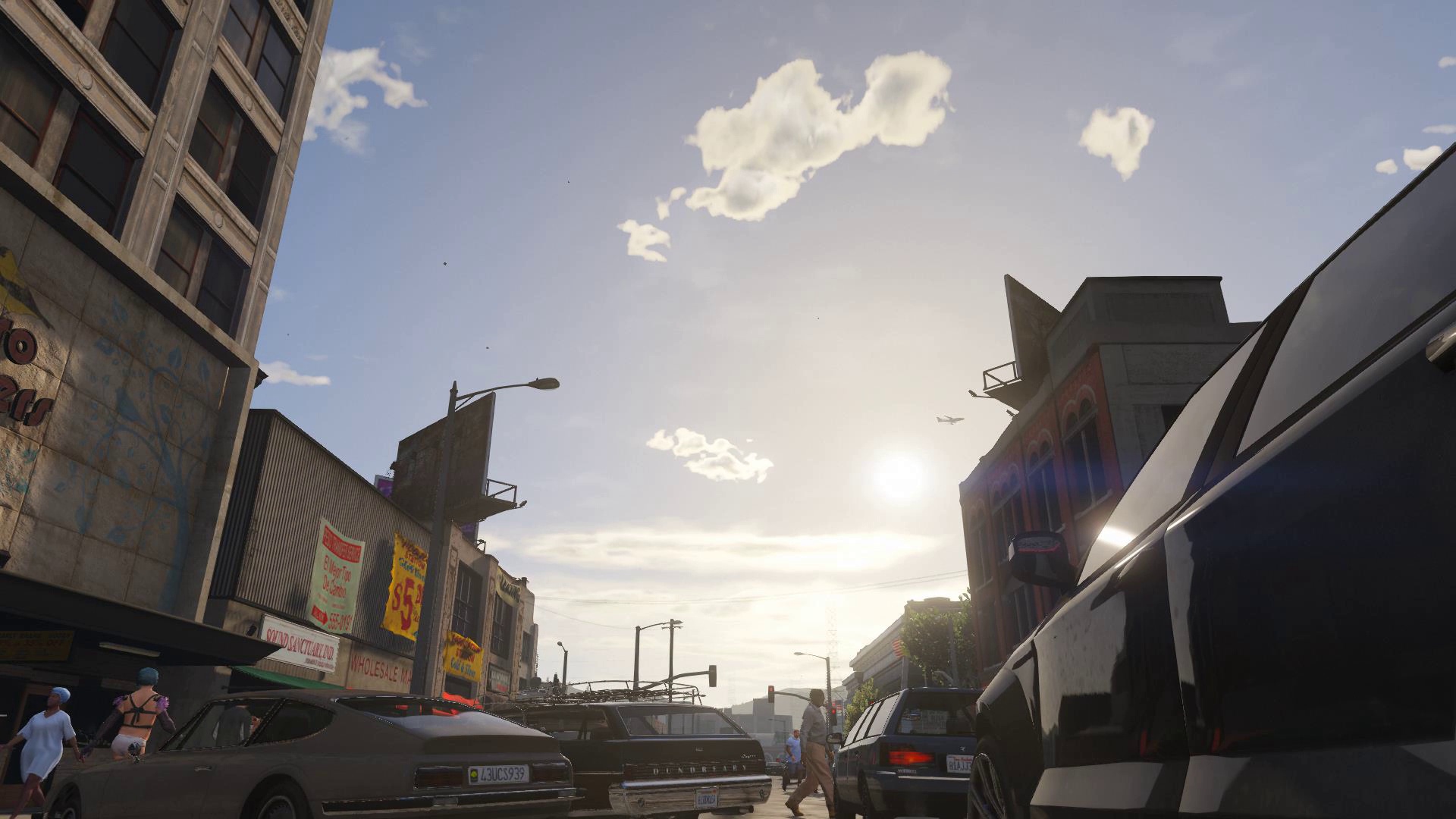}
         \includegraphics[width=\textwidth]{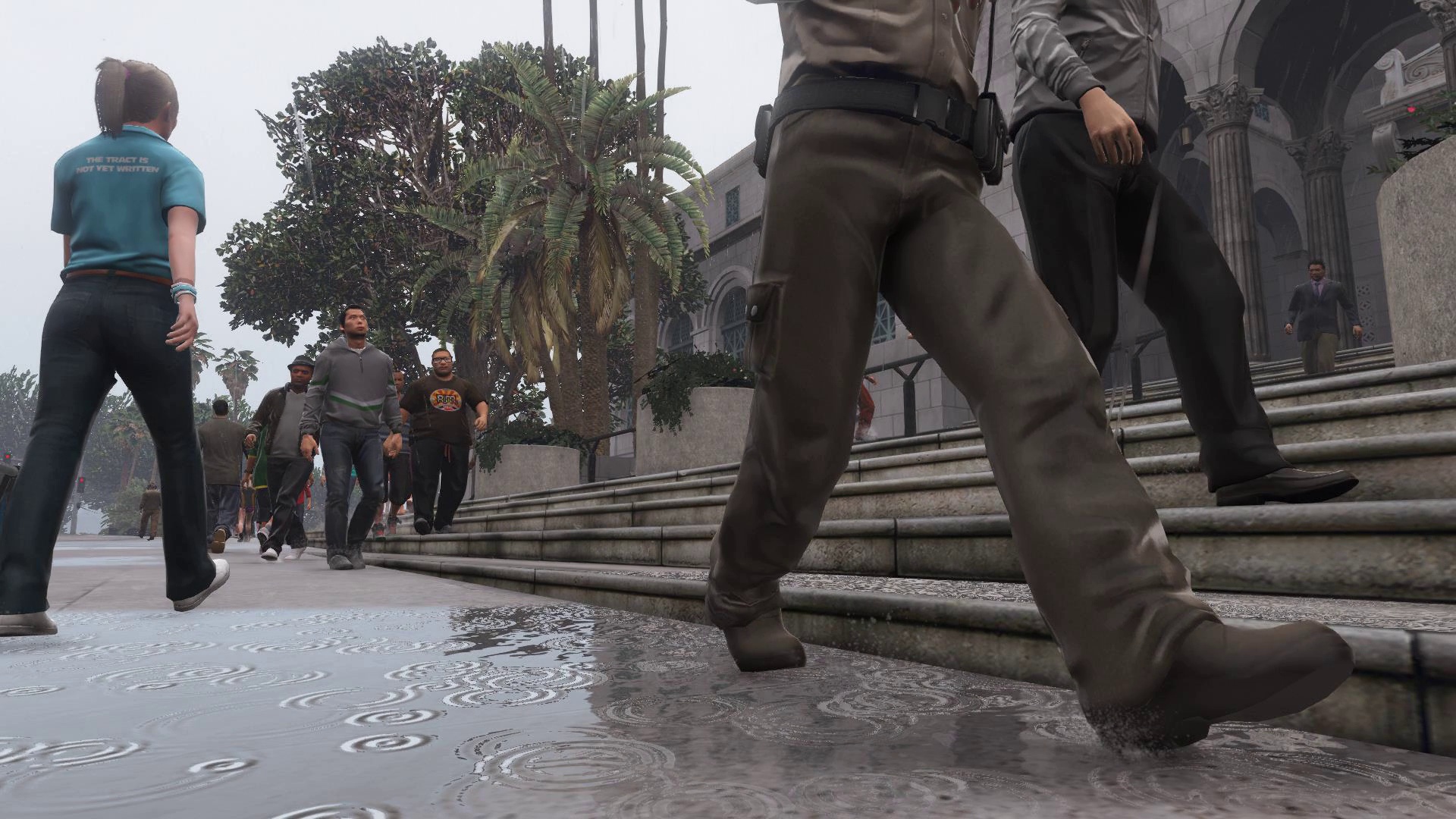}
         \includegraphics[width=\textwidth]{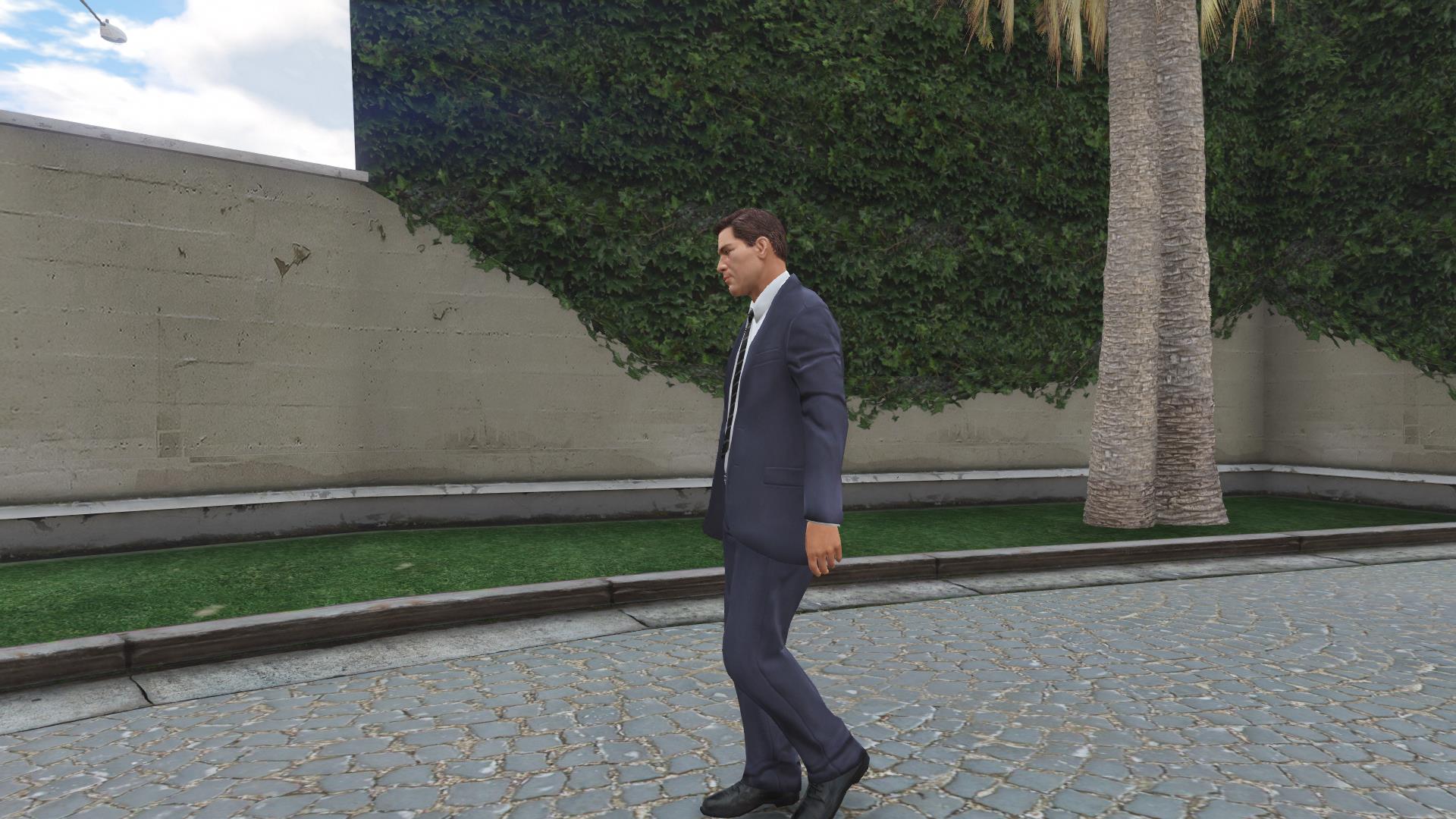}
     \end{subfigure}
     \begin{subfigure}[b]{0.19\textwidth}
         \centering
         \includegraphics[width=\textwidth]{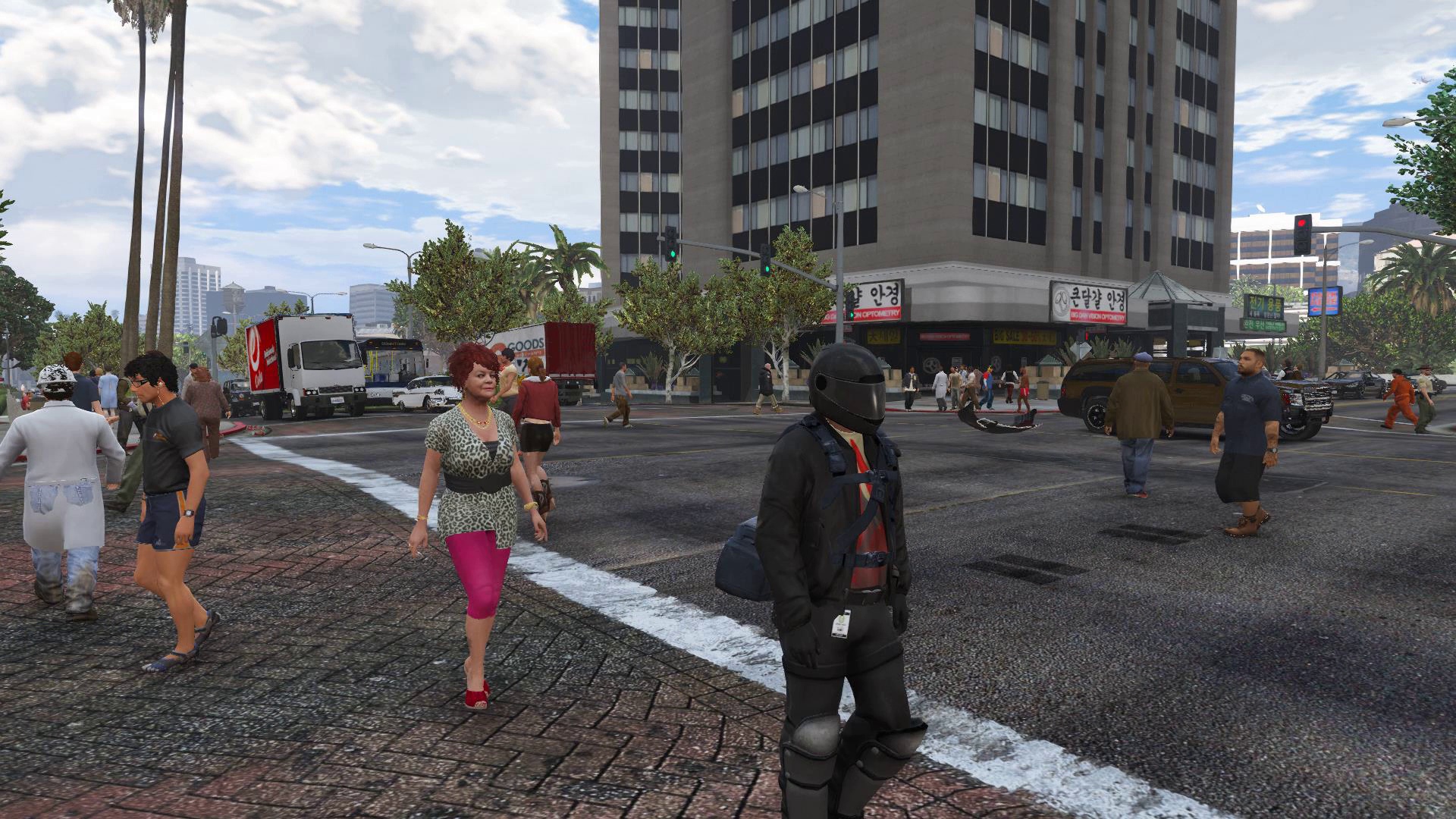}
         \includegraphics[width=\textwidth]{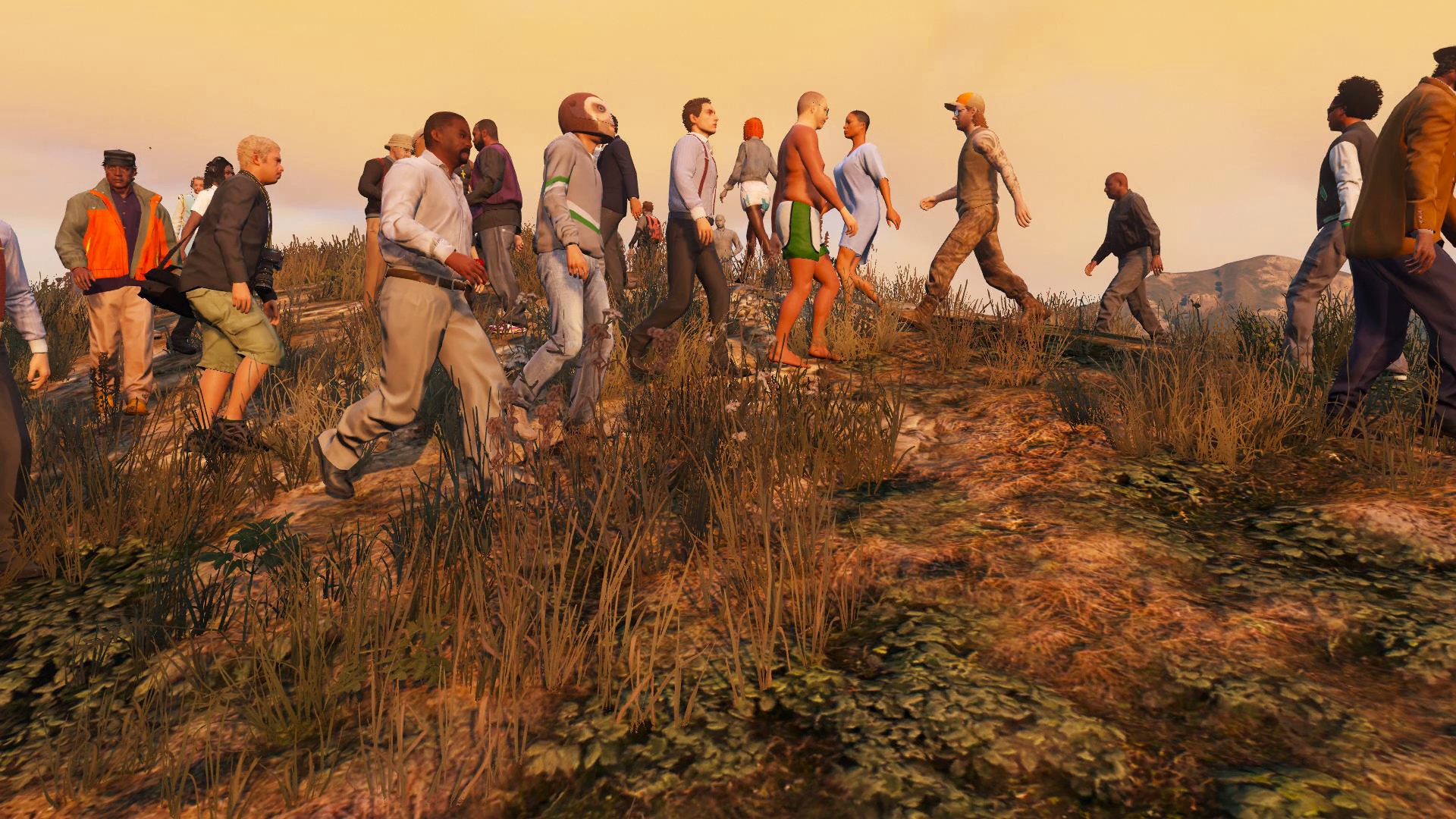}
         \includegraphics[width=\textwidth]{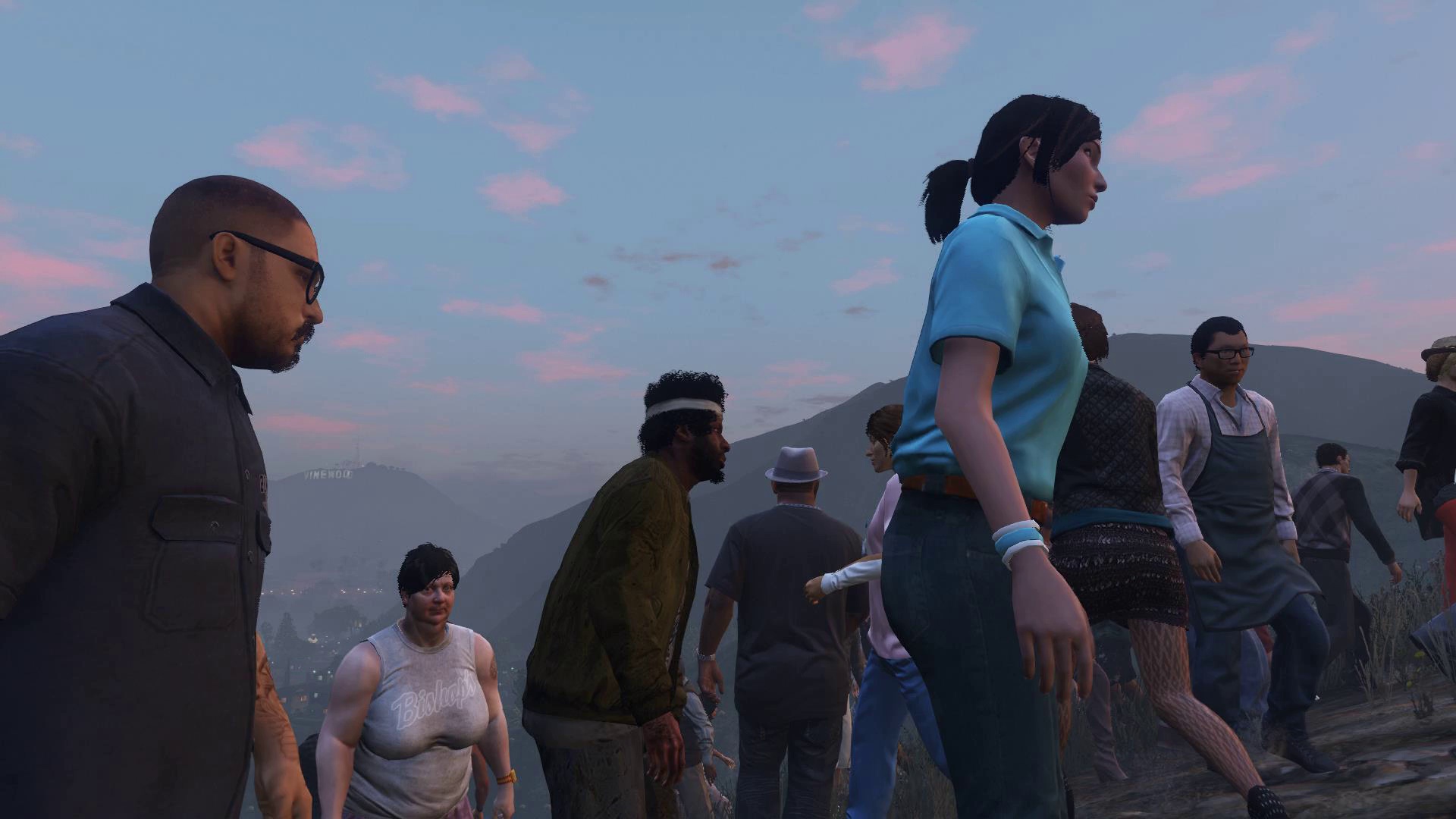}
         \includegraphics[width=\textwidth]{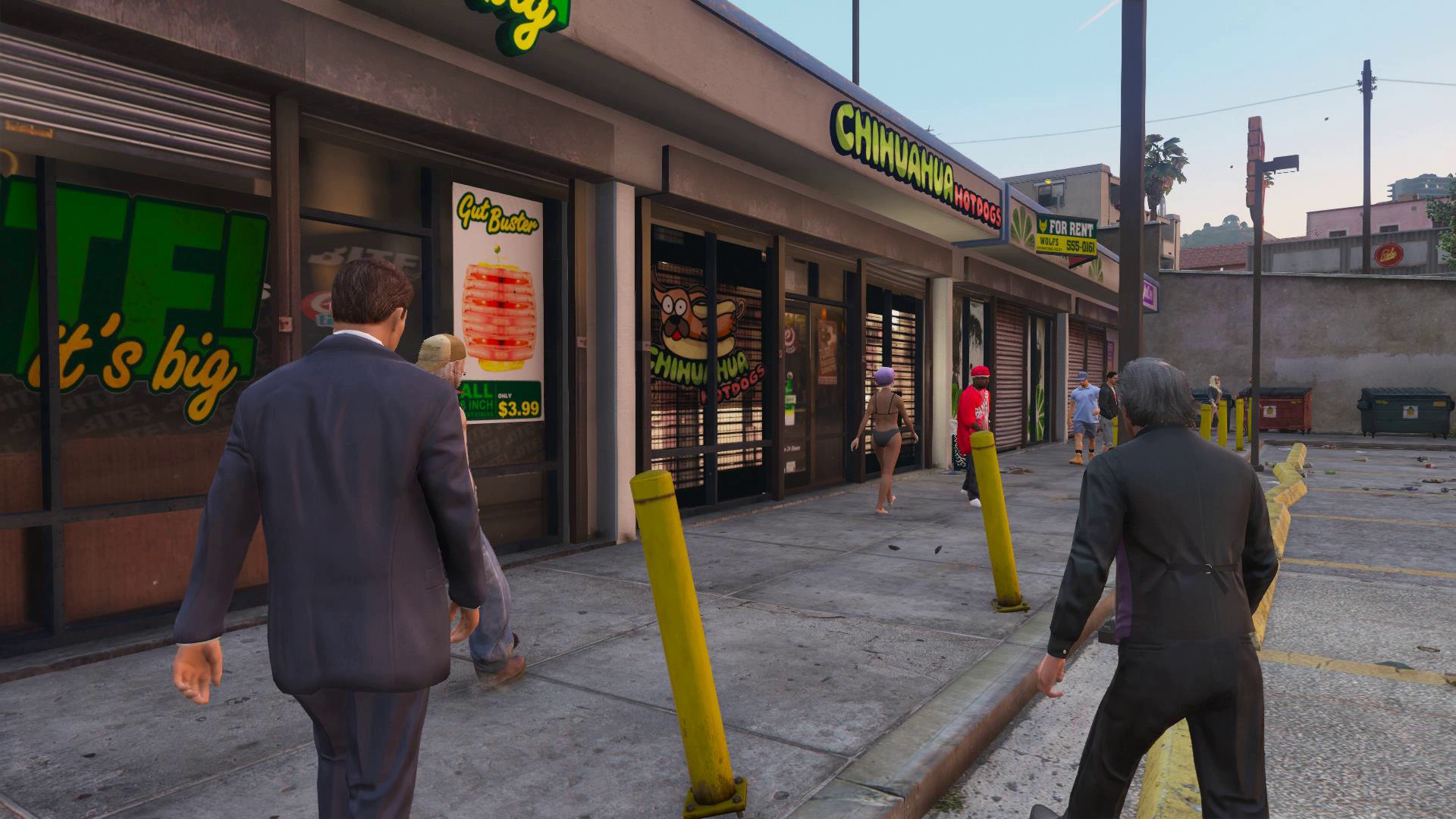}
     \end{subfigure}
     \begin{subfigure}[b]{0.19\textwidth}
         \centering
         \includegraphics[width=\textwidth]{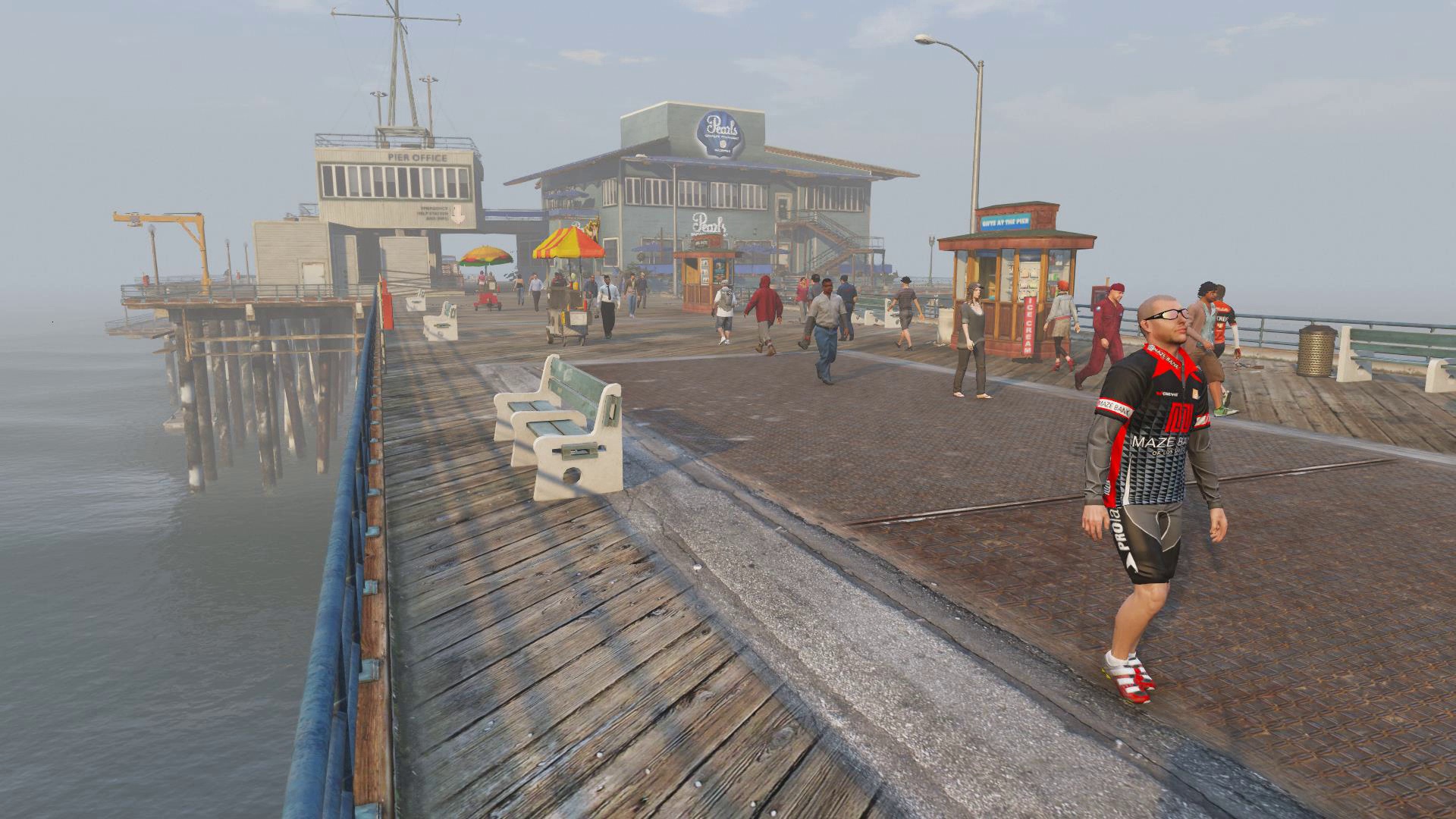}
         \includegraphics[width=\textwidth]{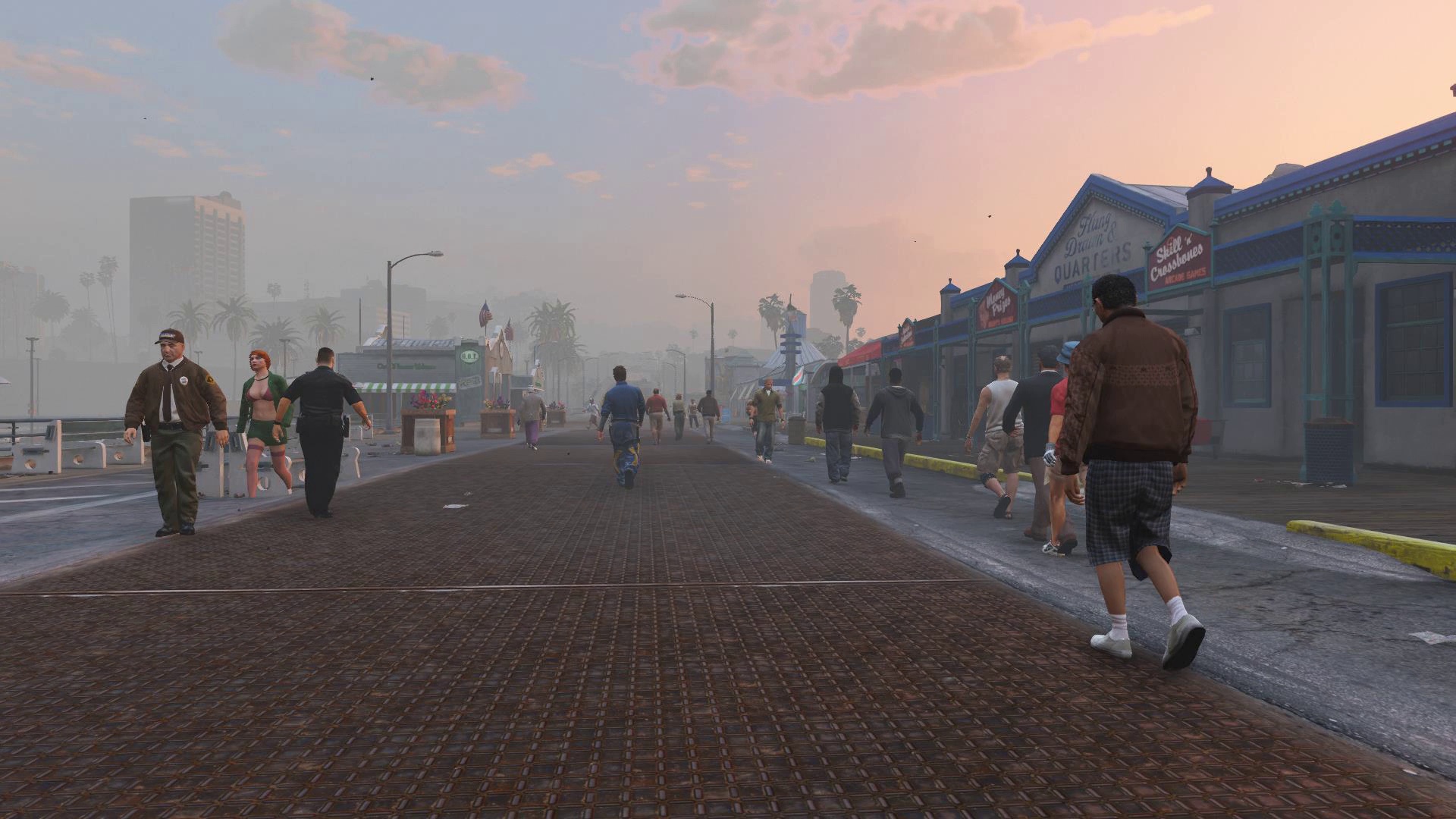}
         \includegraphics[width=\textwidth]{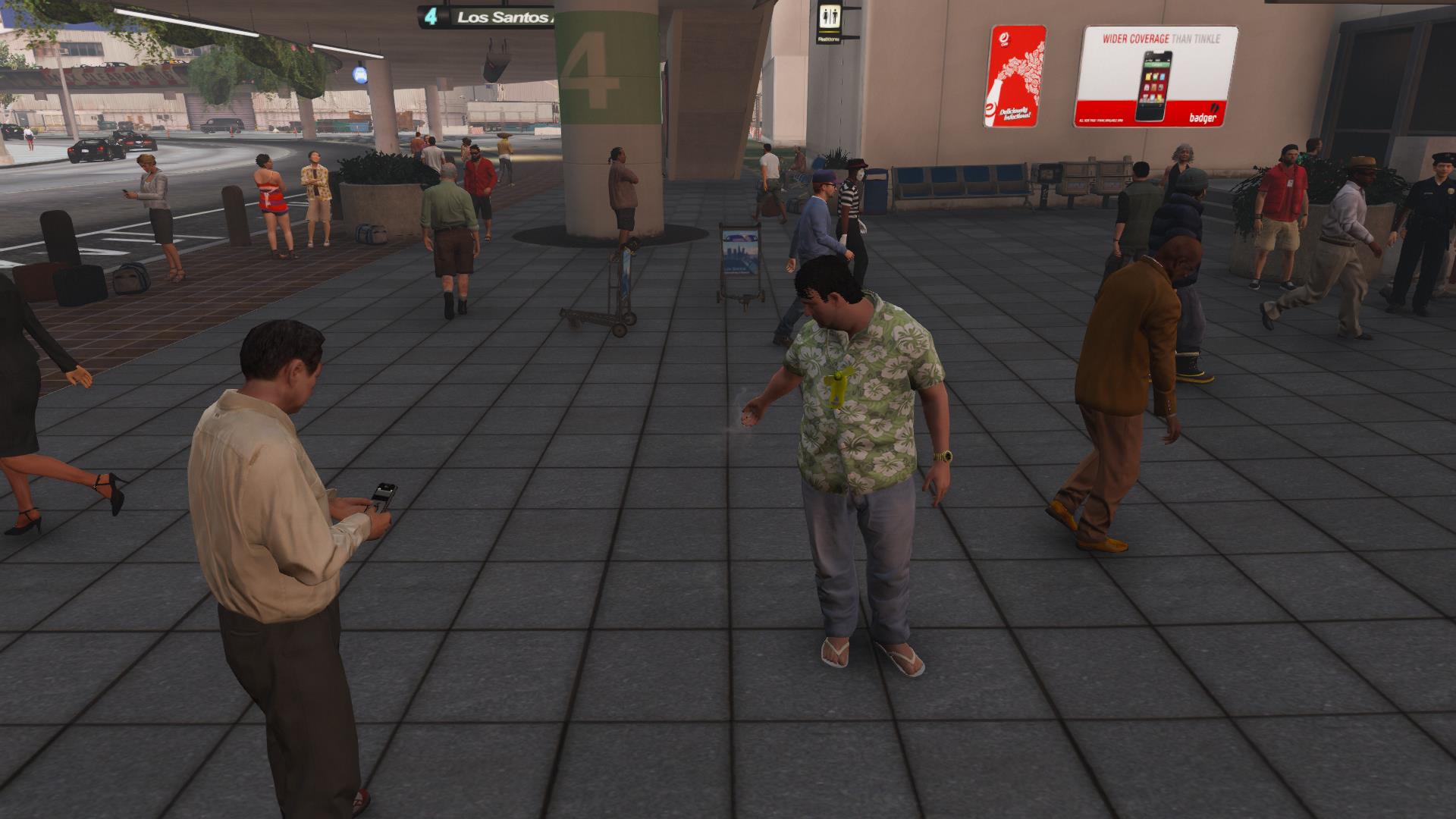}
         \includegraphics[width=\textwidth]{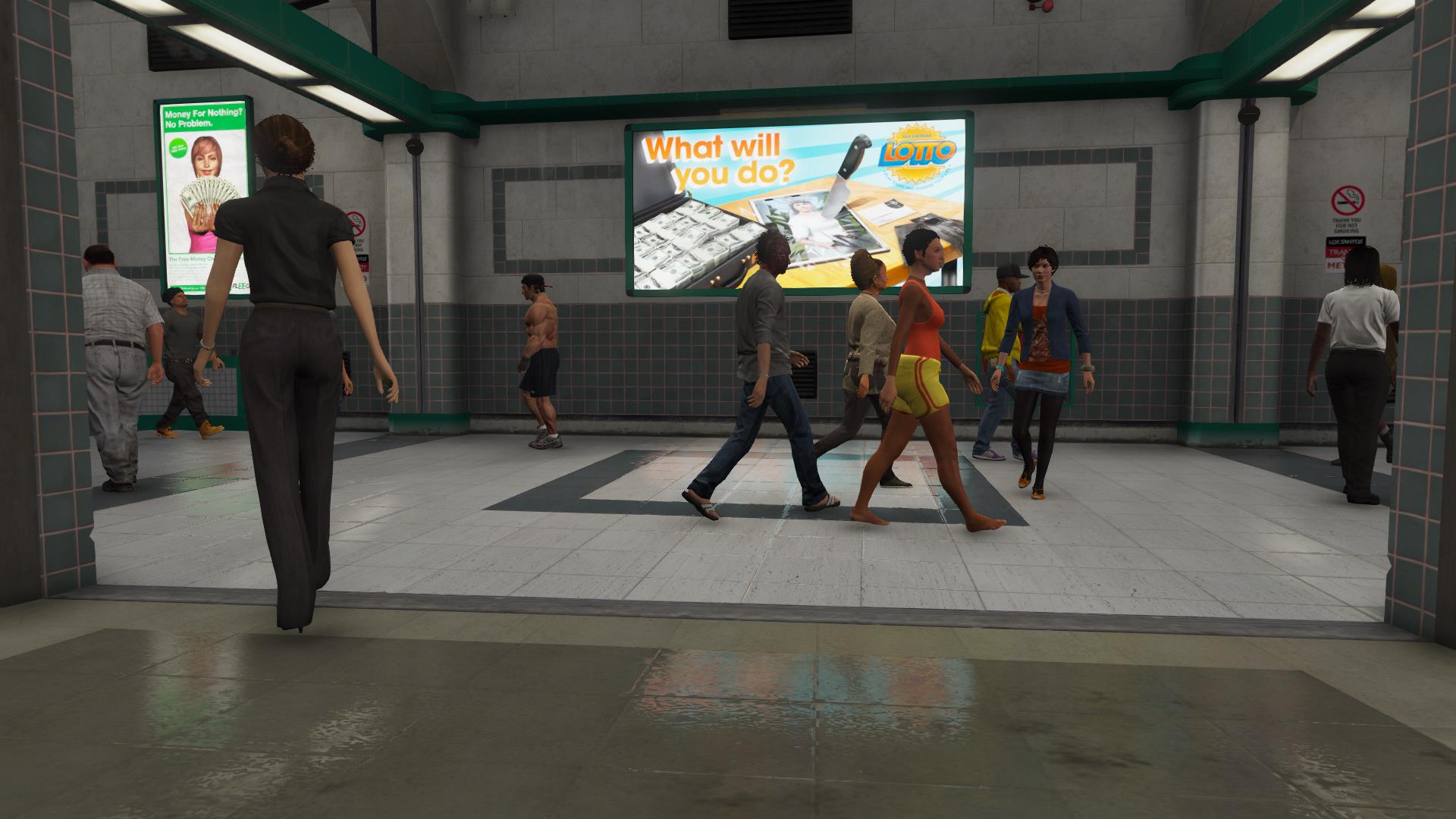}
     \end{subfigure}
     \begin{subfigure}[b]{0.19\textwidth}
         \centering
         \includegraphics[width=\textwidth]{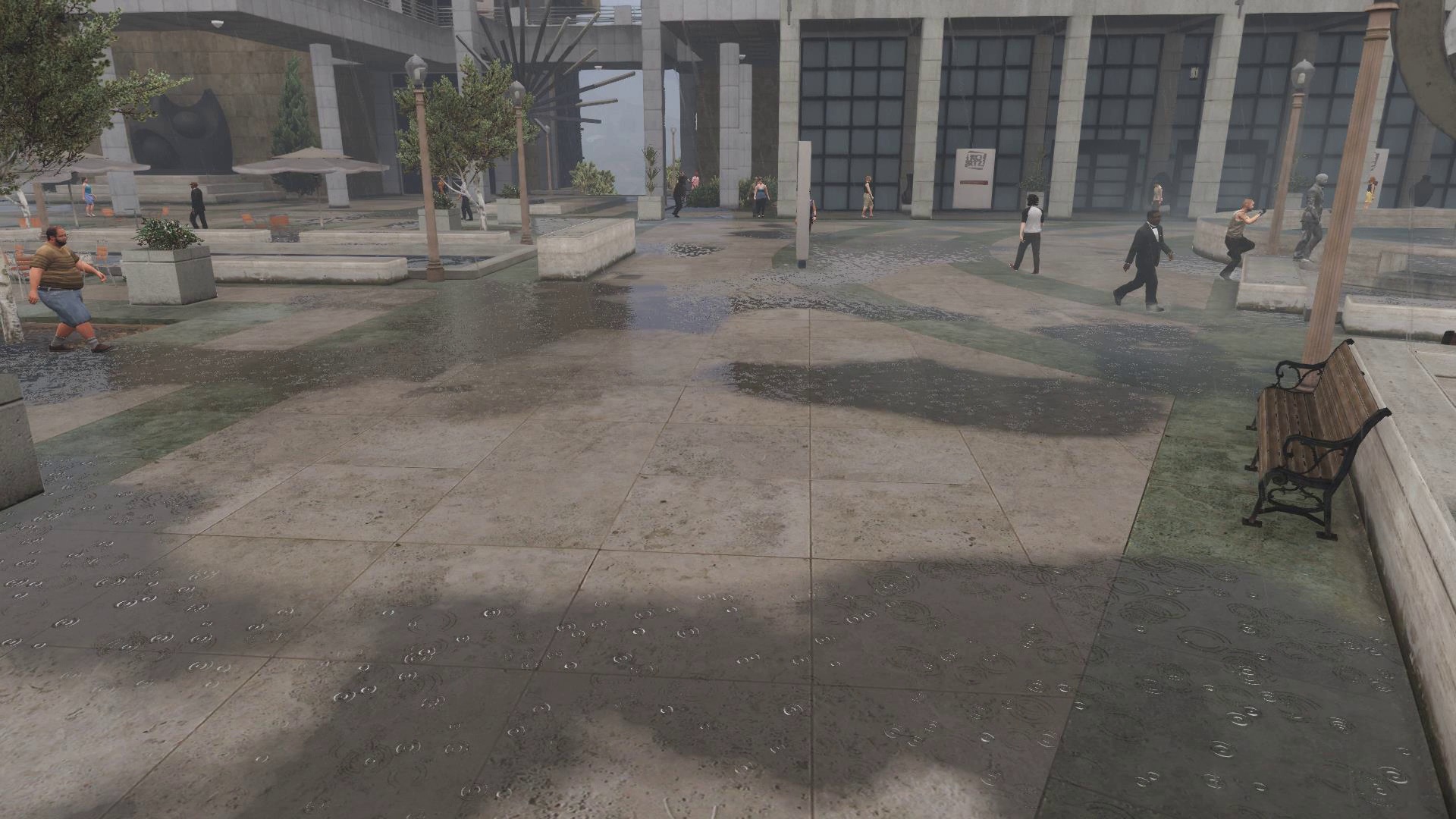}
         \includegraphics[width=\textwidth]{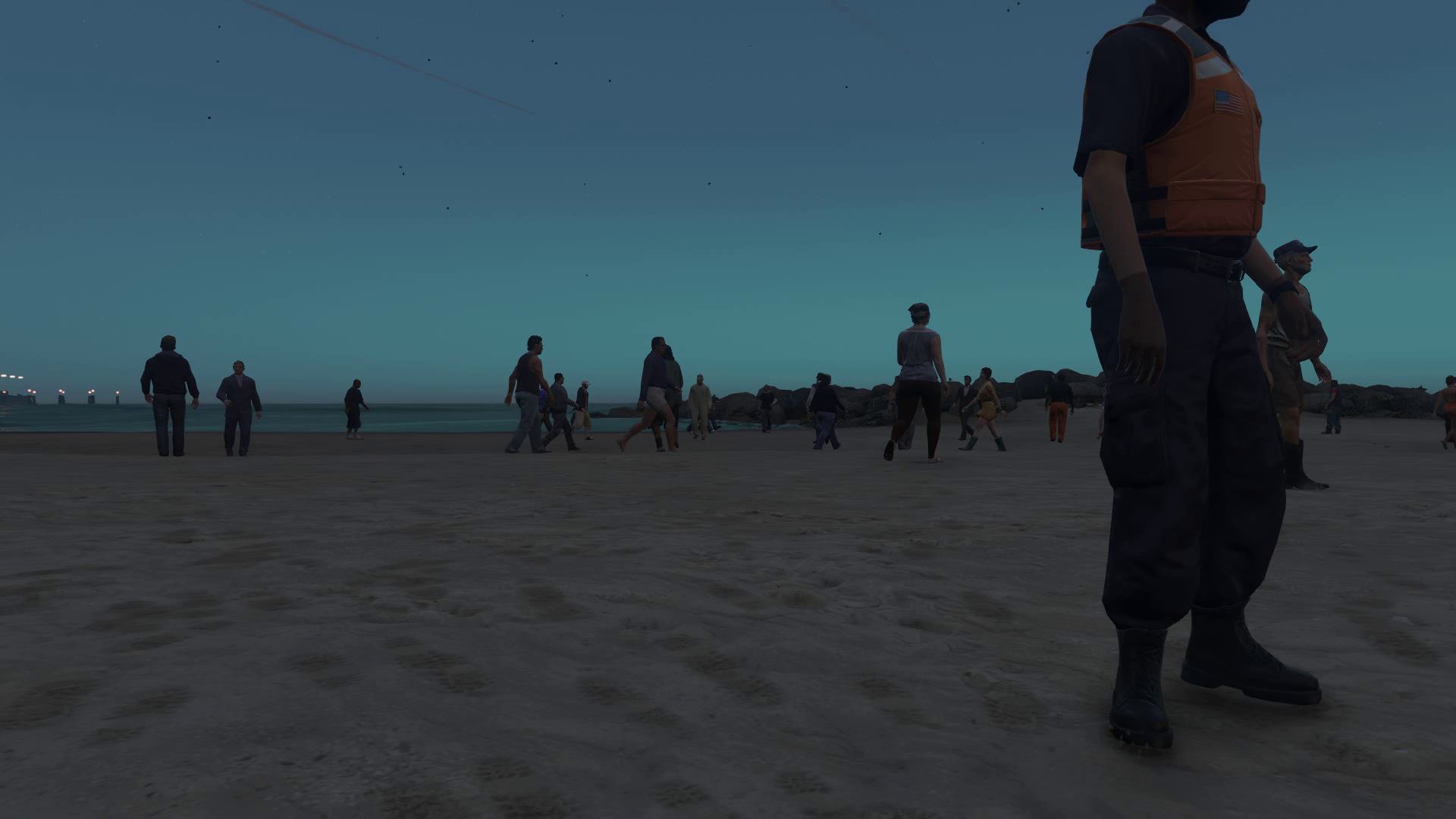}
         \includegraphics[width=\textwidth]{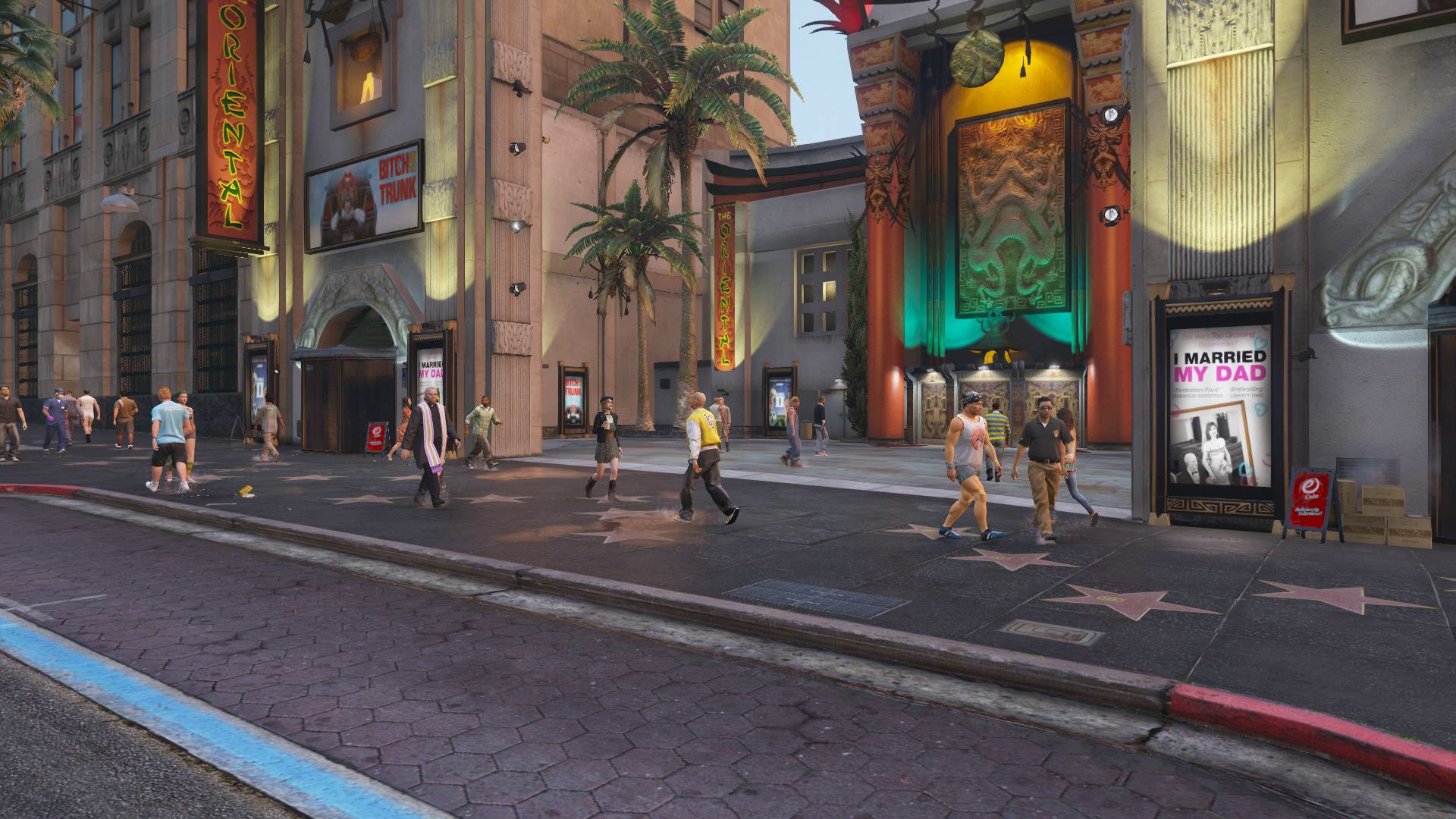}
         \includegraphics[width=\textwidth]{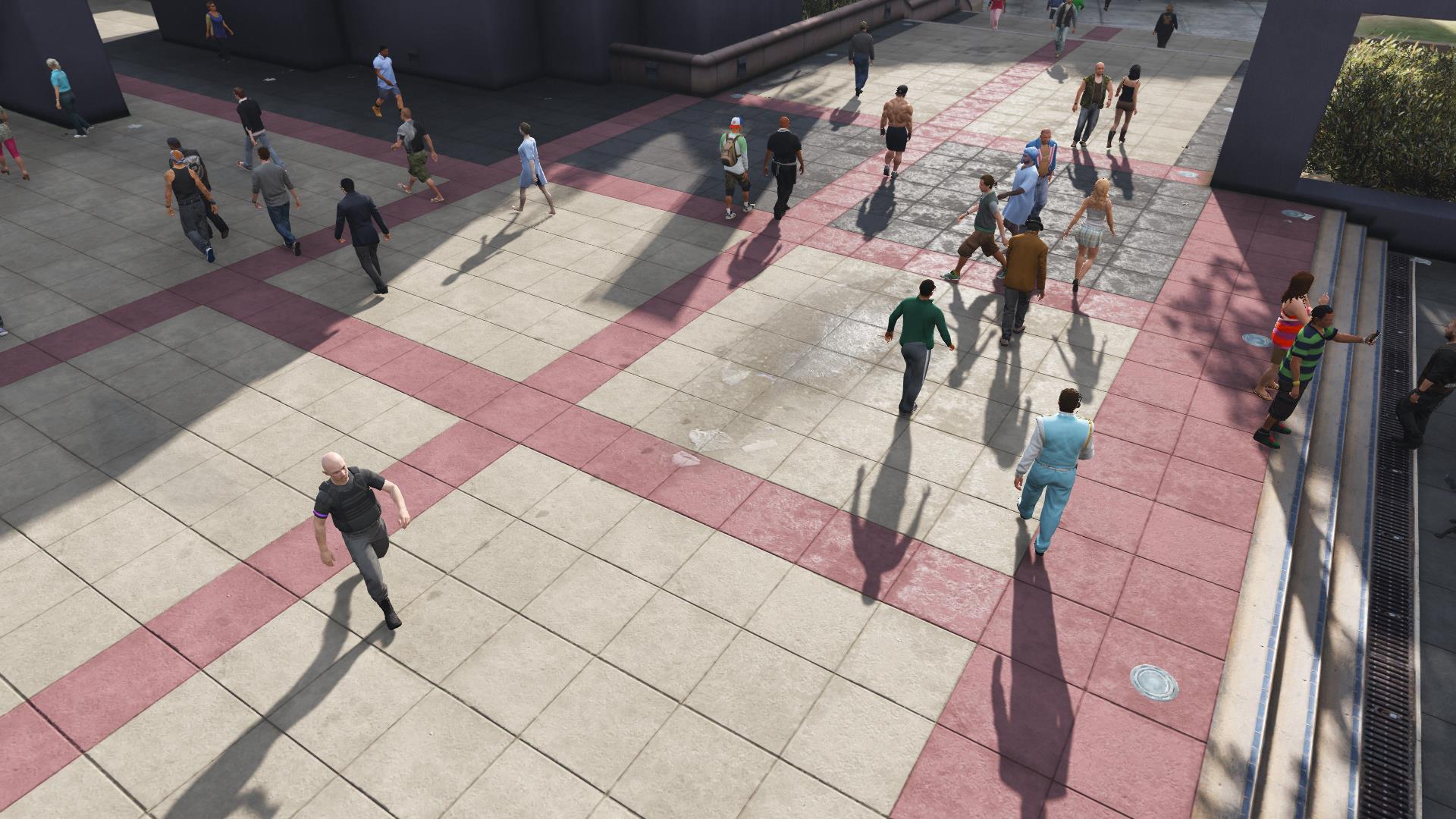}
     \end{subfigure}
     \begin{subfigure}[b]{0.19\textwidth}
         \centering
         \includegraphics[width=\textwidth]{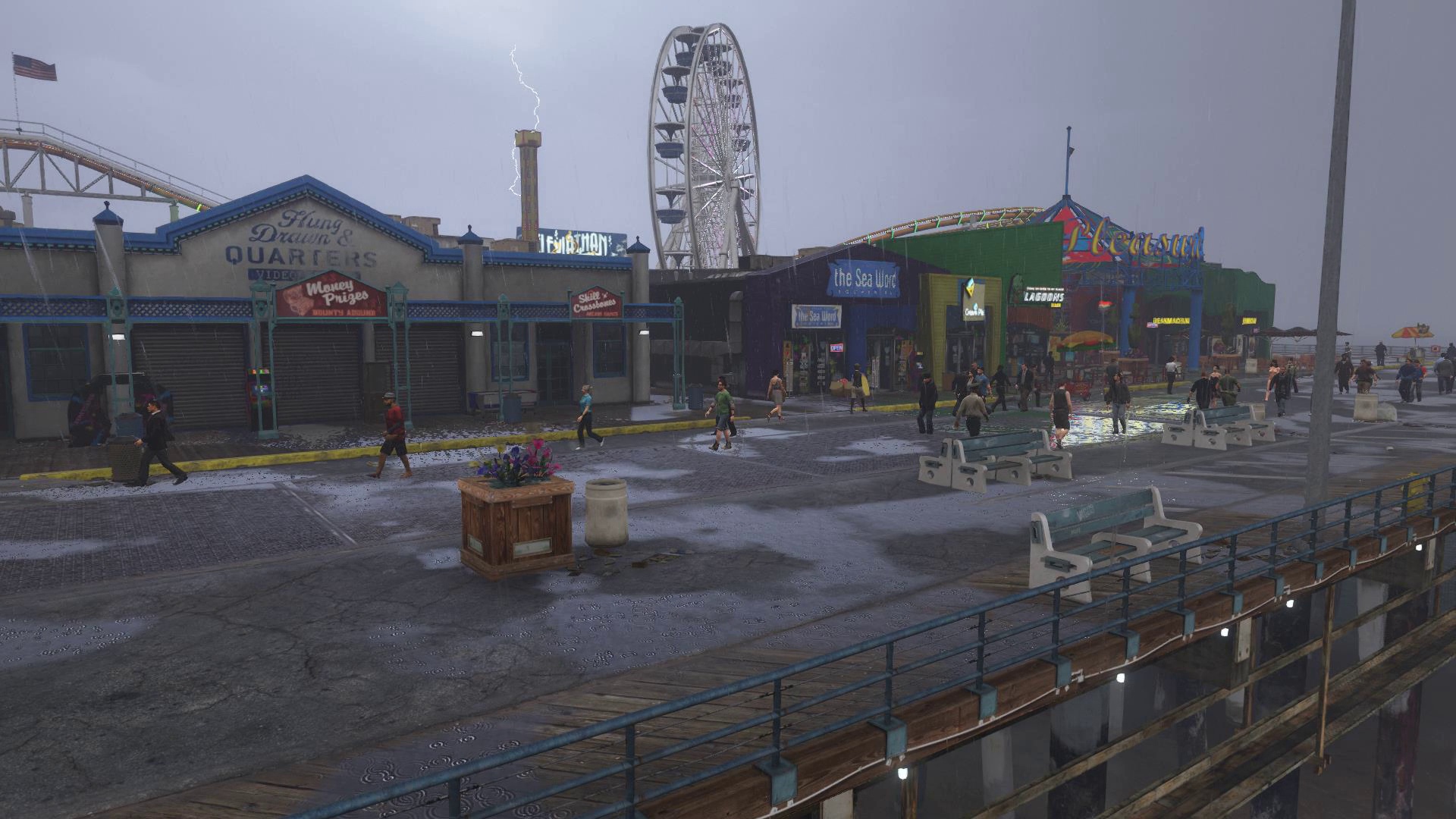}
         \includegraphics[width=\textwidth]{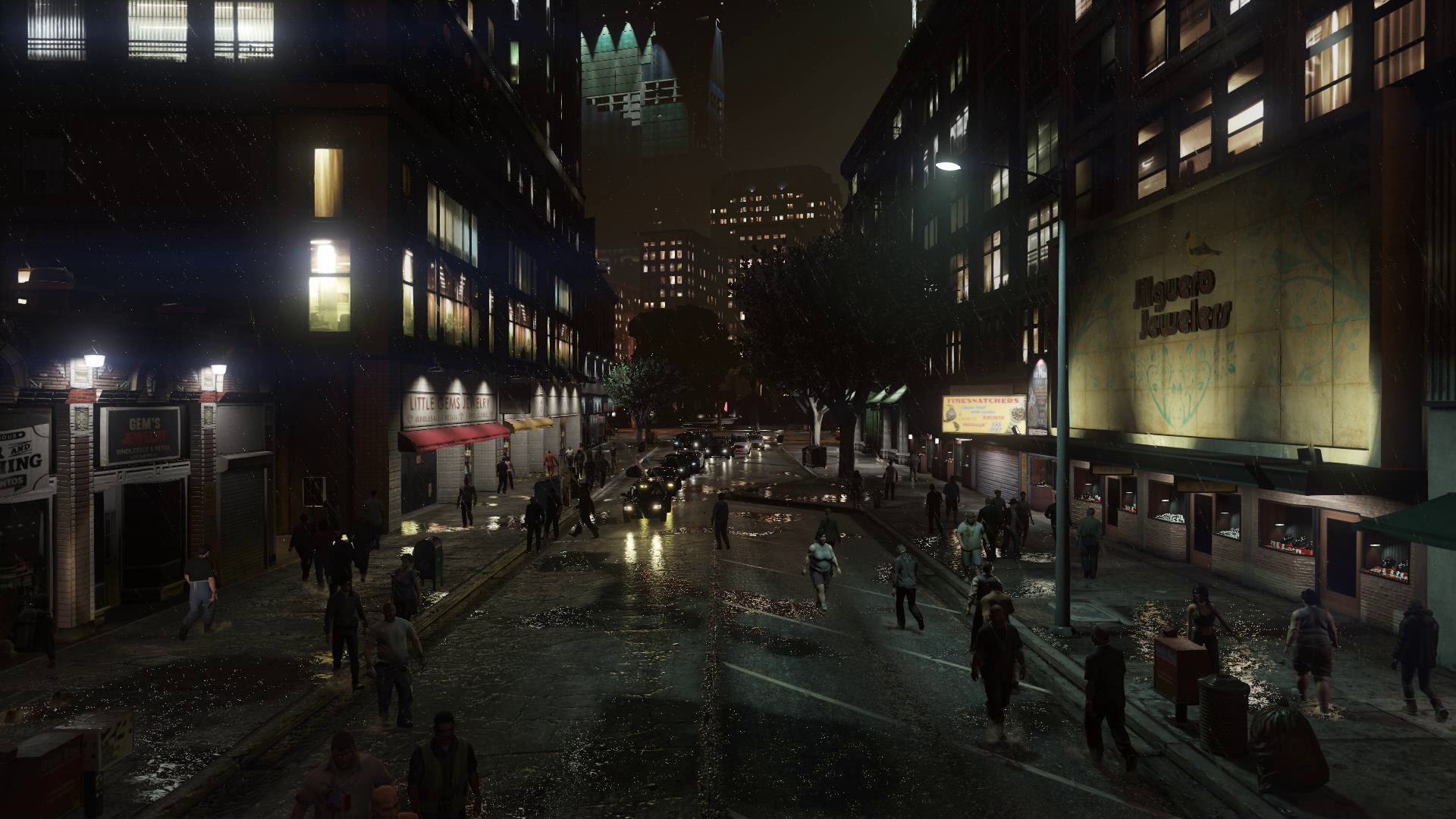}
         \includegraphics[width=\textwidth]{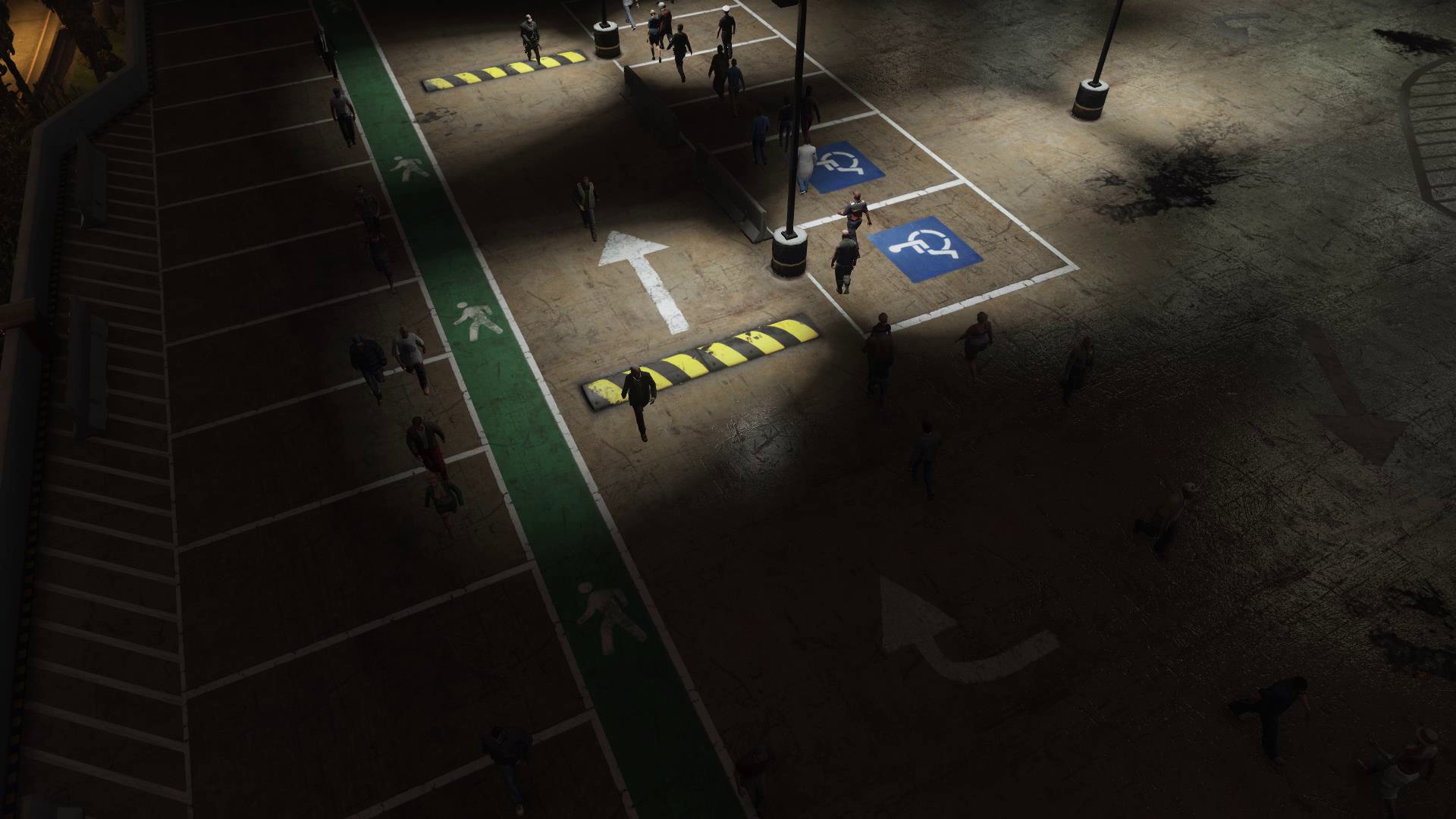}
         \includegraphics[width=\textwidth]{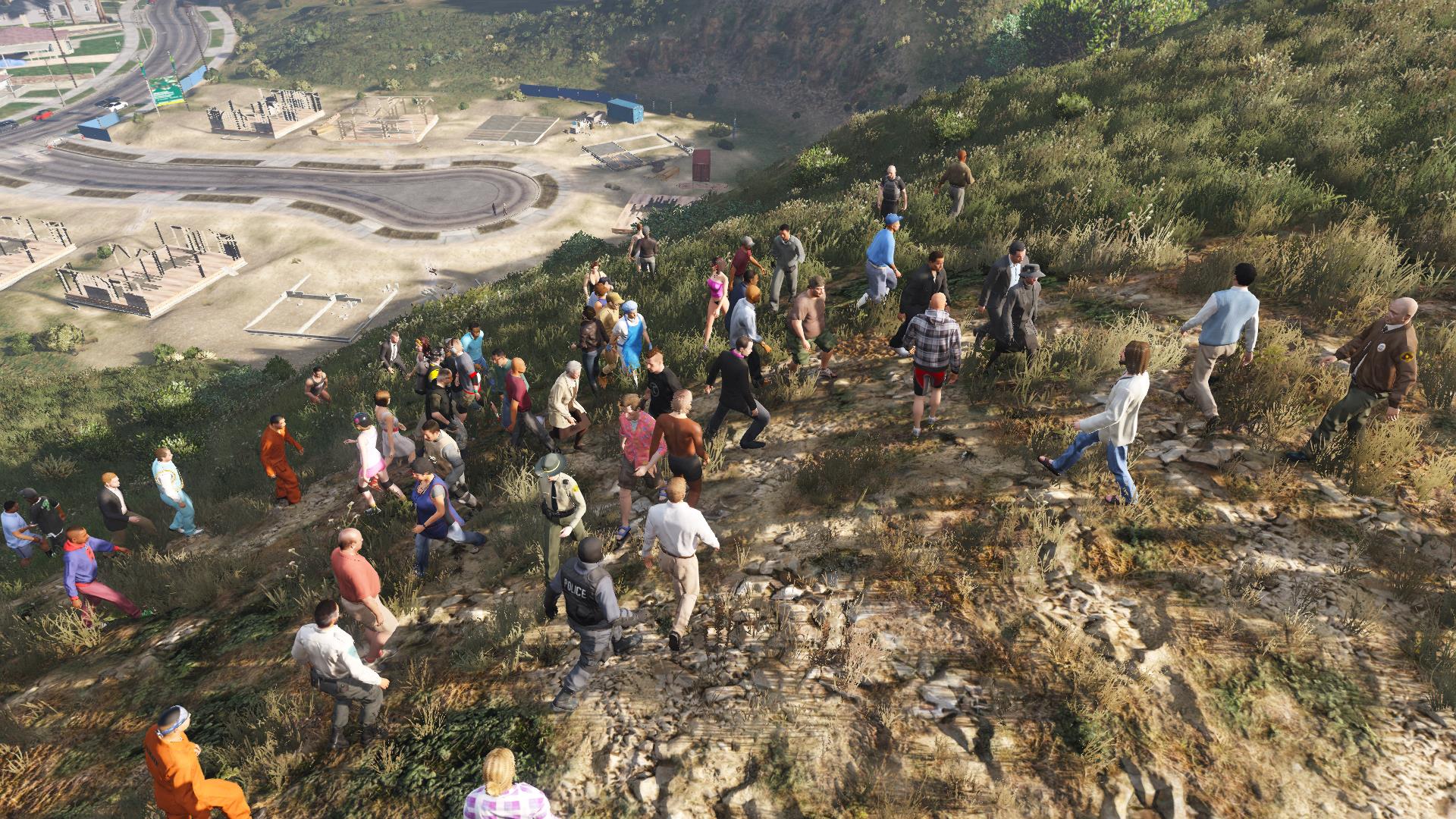}
     \end{subfigure}
\\[1px]
\caption{Examples from the \motsynth dataset showing data variety in terms of weather conditions (first row), lighting condition (second row), viewpoints (third row) and number of people (fourth row). Best viewed on screen.}
\label{fig:variety}
\vspace{0.1cm}
\end{center}
\end{figure*}

\section{Overview}

In this supplementary, we provide (i) extended version of the Tab. 1 (dataset comparison), provided in the main paper (Sec.~\ref{sec:dataset});
(ii) additional dataset visualizations and statistics (Sec.~\ref{sec:dataset-extra}); (iii) additional experiments on trade-offs on data volume vs. diversity (Sec.~\ref{sec:volume-diversity}); (iv) implementation details for all experiments, provided in the main paper (Sec.~\ref{sec:implementation}); (v)
MOT20 benchmark results for each sequence (Sec.~\ref{sec:mot20_detailed}).

\section{Dataset Comparison}
\label{sec:dataset}

Tab.~\ref{table:datasets} extends Tab. 1 from the main paper. In the table the most widely used publicly available datasets that contain annotation for the people class are reported. Compared to real world urban surveillance dataset, \motsynth has one order of magnitude more clips, annotated frames and annotated instances. Besides JTA~\cite{fabri18eccv}, \motsynth is the only available dataset that provides 3D pose annotations. Additionally, \motsynth also provides instance segmentation labels and depth maps. It is important to note that for autonomous driving datasets \cite{Geiger12CVPR,caesar2020nuscenes,Yuan183DV,sun20CVPR} and TAO~\cite{dave20eccv} the number of instances is relative to all the classes;

\begin{table}[ht]
\centering
\resizebox{1.0\linewidth}{!}{
\begin{tabular}{c|l|c|c|r|c|c}
\toprule

& Dataset  & Split & Sampling rate & frames & AP & MODA \\
\midrule
\parbox[t]{3mm}{\multirow{9}{*}{\rotatebox[origin=c]{90}{YOLOv3}}}
& COCO & -- & -- & 118k & 69.76 & 62.02 \\
\cline{2-7}
& \multirow{7}{*}{\motsynth} 
 & \multirow{2}{*}{1} & 1:60 & 2k & 51.15 & 45.71 \\
&  &  & 1:10 & 13k & 62.66 & 52.36 \\
 \cline{3-7}
&  & \multirow{2}{*}{2} & 1:60 & 4k & 53.86 & 47.49 \\
&  &  & 1:10 & 24k & 63.08 & 56.67 \\
  \cline{3-7}
& & \multirow{2}{*}{3} & 1:60 & 9k & 62.10 & 51.20 \\
& &  & 1:10 & 52k & 63.13 & 60.60 \\
 \cline{3-7}
& & \multirow{2}{*}{4} & 1:60 & 17k & 62.59 & 58.66 \\
& &  & 1:10 & 104k & \textbf{71.90} & \textbf{64.51} \\
\midrule
\parbox[t]{3mm}{\multirow{9}{*}{\rotatebox[origin=c]{90}{CenterNet}}}
& COCO & -- & -- & 118k & 67.01 & 44.38 \\
\cline{2-7}
& \multirow{7}{*}{\motsynth} 
 & \multirow{2}{*}{1} & 1:60 & 2k & 61.18 & 39.06 \\
&  &  & 1:10 & 13k& 61.82 & 49.34 \\
 \cline{3-7}
&  & \multirow{2}{*}{2} & 1:60 & 4k  & 61.45 & 44.54 \\
&  &  & 1:10 & 24k & 62.32 & 54.90 \\
  \cline{3-7}
& & \multirow{2}{*}{3} & 1:60 & 9k & 62.22 & 53.04 \\
& &  & 1:10 & 52k & 62.45 & 55.82 \\
 \cline{3-7}
& & \multirow{2}{*}{4} & 1:60 & 17k & 70.15 & 51.75 \\
& &  & 1:10 & 104k & \textbf{70.68} & \textbf{57.39} \\
\midrule
\parbox[t]{3mm}{\multirow{9}{*}{\rotatebox[origin=c]{90}{ Faster R-CNN}}}
& COCO & -- & -- & 118k & 76.68 & 53.86 \\
\cline{2-7}
& \multirow{7}{*}{\motsynth} 
 & \multirow{2}{*}{1} & 1:60 & 2k & 70.00 & 42.90 \\
&  &  & 1:10 & 13k & 76.80 & 39.02 \\
 \cline{3-7}
&  & \multirow{2}{*}{2} & 1:60 & 4k & 70.27 & 44.54 \\
&  &  & 1:10 & 24k & 77.47 & 50.62 \\
  \cline{3-7}
& & \multirow{2}{*}{3} & 1:60 & 9k & 77.32 & 51.46 \\
& &  & 1:10 & 52k & 78.30 & 49.75 \\
 \cline{3-7}
& & \multirow{2}{*}{4} & 1:60 & 17k & 77.78 & 53.72 \\
& &  & 1:10 & 194k & \textbf{78.98} & \textbf{54.96} \\
\bottomrule
\end{tabular}
}
\caption{The effect of the density of sampled data. Sparser sampling increases the diversity. As can be seen, we can bridge the gap syn-to-real even when using smaller \motsynth subsets if we ensure that training images are diverse.}
\label{tab:frcnn_fps}
\end{table}

\section{Dataset Visualizatons and Statistics}
\label{sec:dataset-extra}

In Fig.~\ref{fig:variety} we show examples from the \motsynth dataset to demonstrate its variation in terms of weather conditions, lighting conditions, viewpoints, and pedestrian density. 
We recorded sequences exhibiting nine different types of weather: \textit{clear, extra sunny, cloudy, overcast, rainy, thunder, smog, foggy, and blizzard}. 

In addition, \motsynth varies in terms of: (i) \textit{lighting conditions}, resembling different day-time conditions, such as \textit{sunrise, sunset, evening, dawn and night}; (ii) the \textit{camera viewpoint}, ranging from ground plane position to bird's-eye view, and (iii) \textit{density}, ranging from few pedestrians to hundreds of pedestrians. 

\begin{figure*}[ht]
\begin{center}
\vspace{-0.5cm}
     \begin{subfigure}[b]{0.49\textwidth}
         \centering
         \captionsetup{margin=1cm}
         \includegraphics[width=\textwidth]{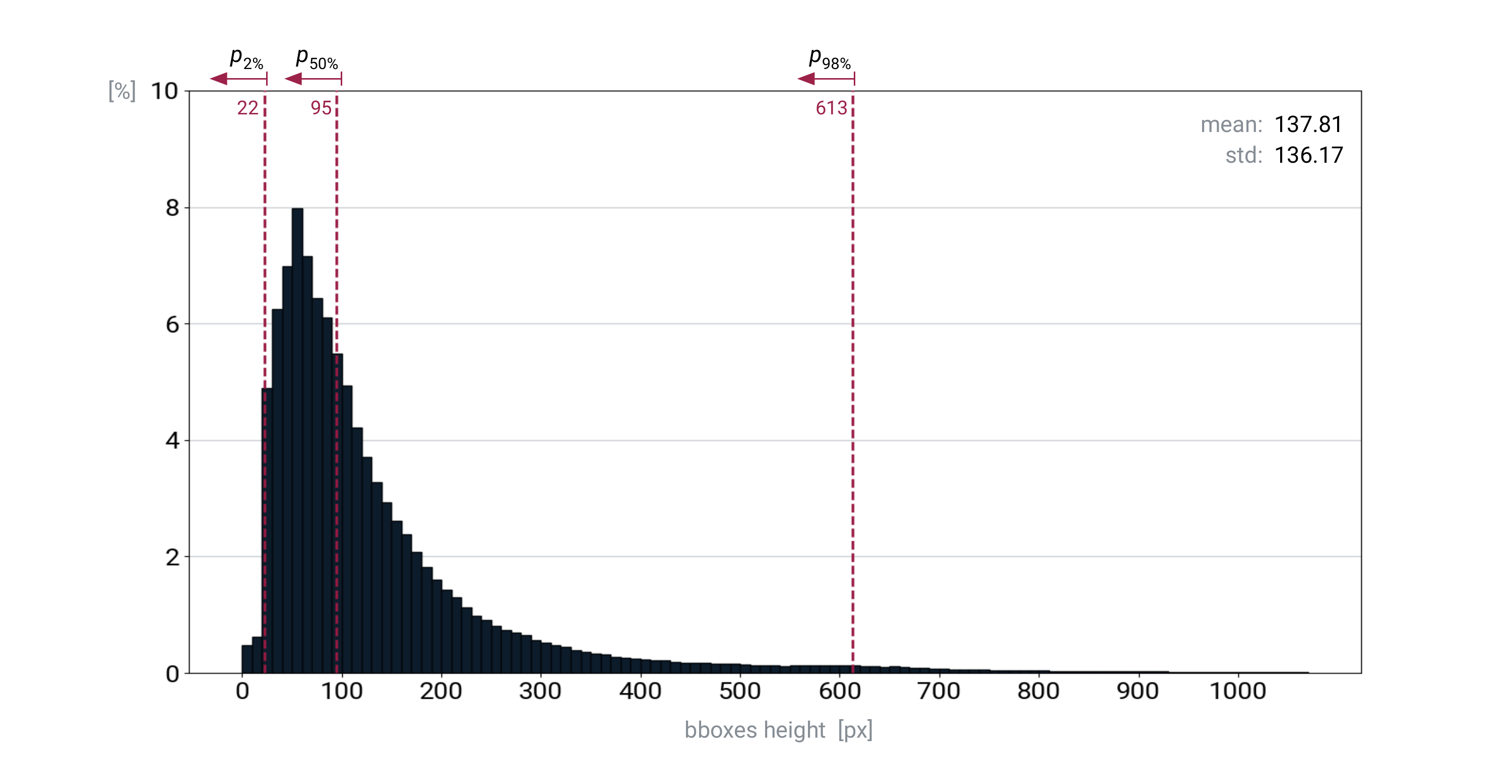}
         \caption{Distribution of the bounding box heights over Full HD images}
         \label{fig:height}
     \end{subfigure}
     \begin{subfigure}[b]{0.49\textwidth}
         \centering
         \captionsetup{margin=1cm}
         \includegraphics[width=\textwidth]{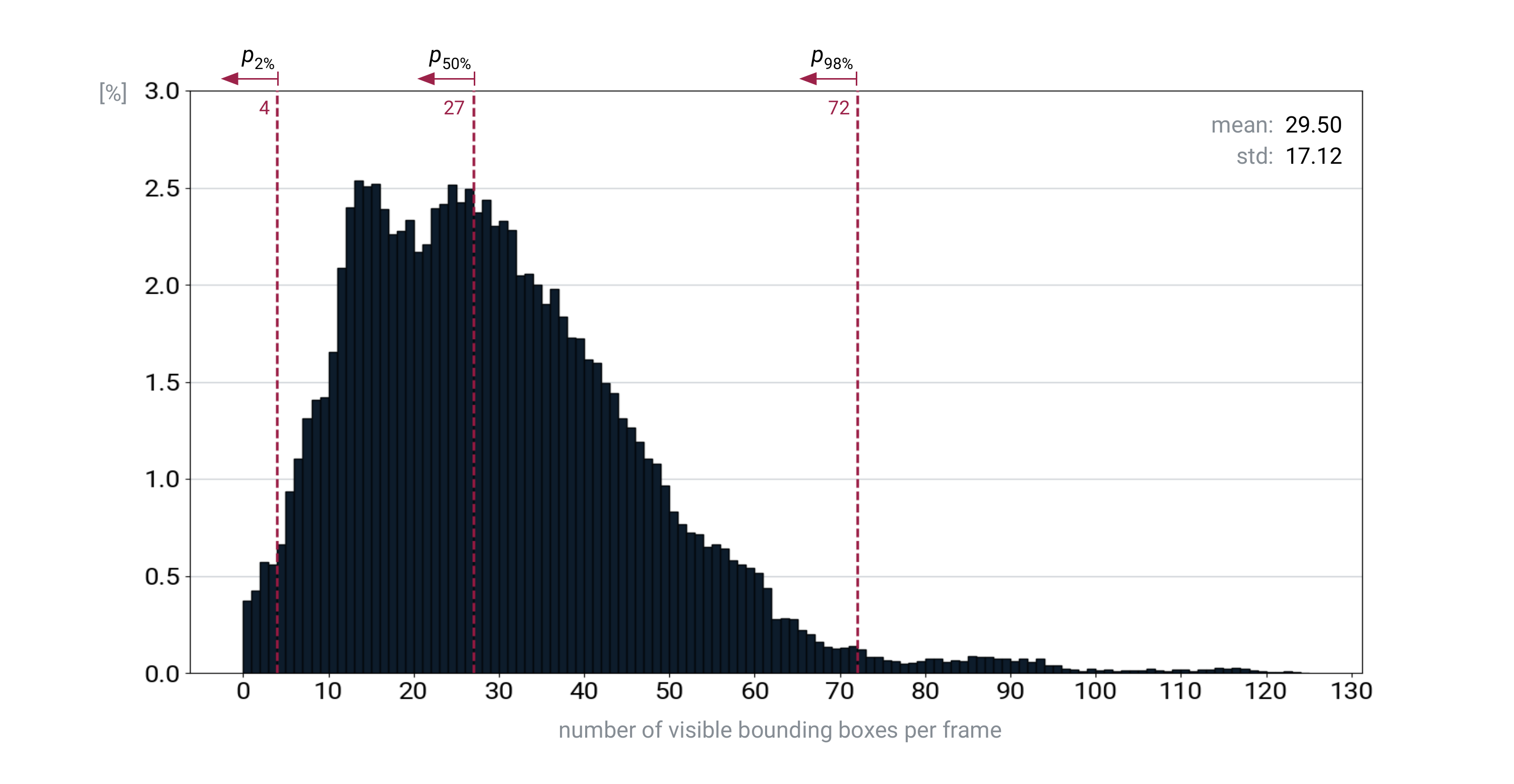}
         \caption{Distribution of the number of bounding boxes per frame.}
         \label{fig:bbox}
     \end{subfigure}
     \begin{subfigure}[b]{0.49\textwidth}
         \centering
         \captionsetup{margin=1cm}
         \includegraphics[width=\textwidth]{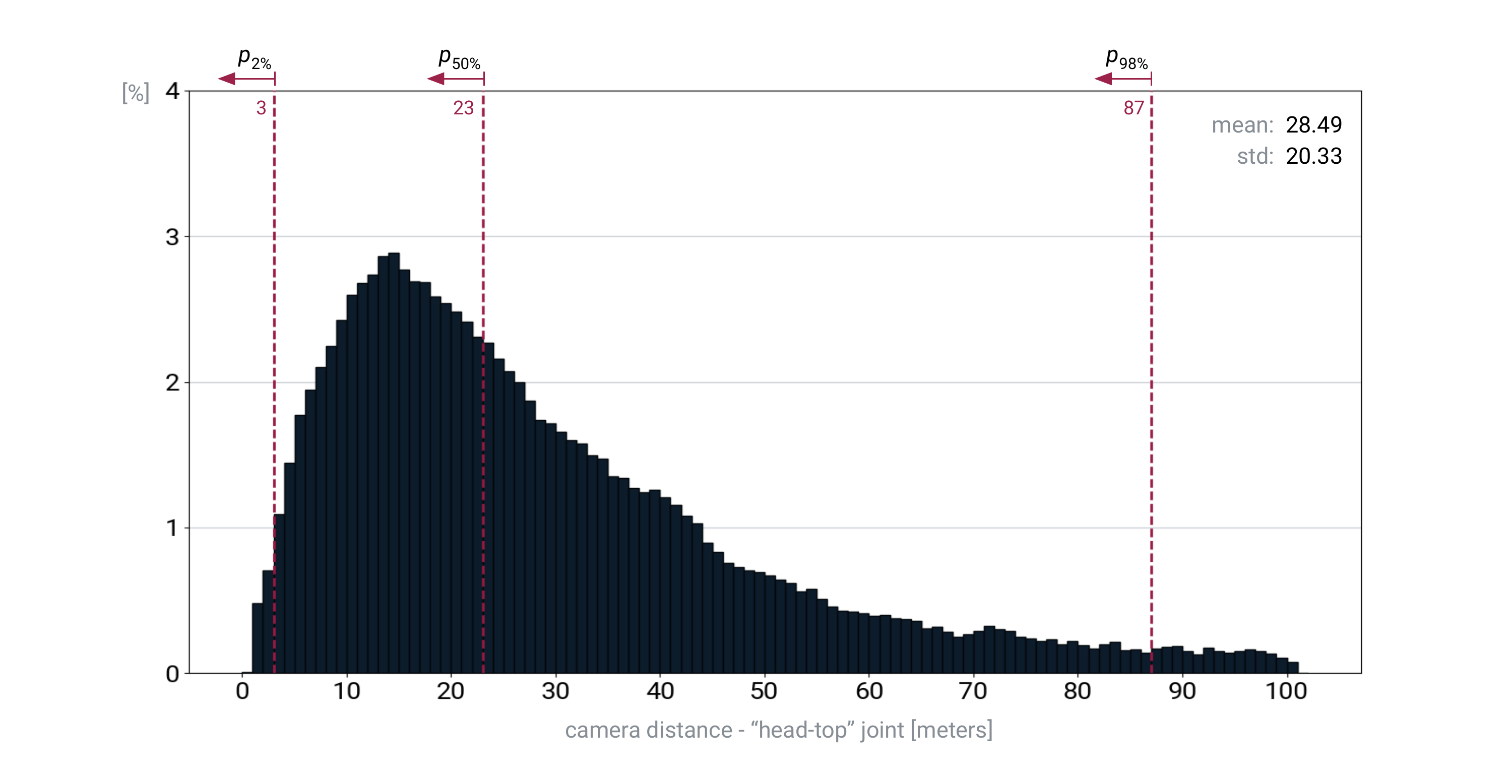}
         \caption{Camera distance distribution of every annotated pedestrian relative to the head joint. }
         \label{fig:distance}
     \end{subfigure}
     \begin{subfigure}[b]{0.49\textwidth}
         \centering
         \captionsetup{margin=1cm}
         \includegraphics[width=\textwidth]{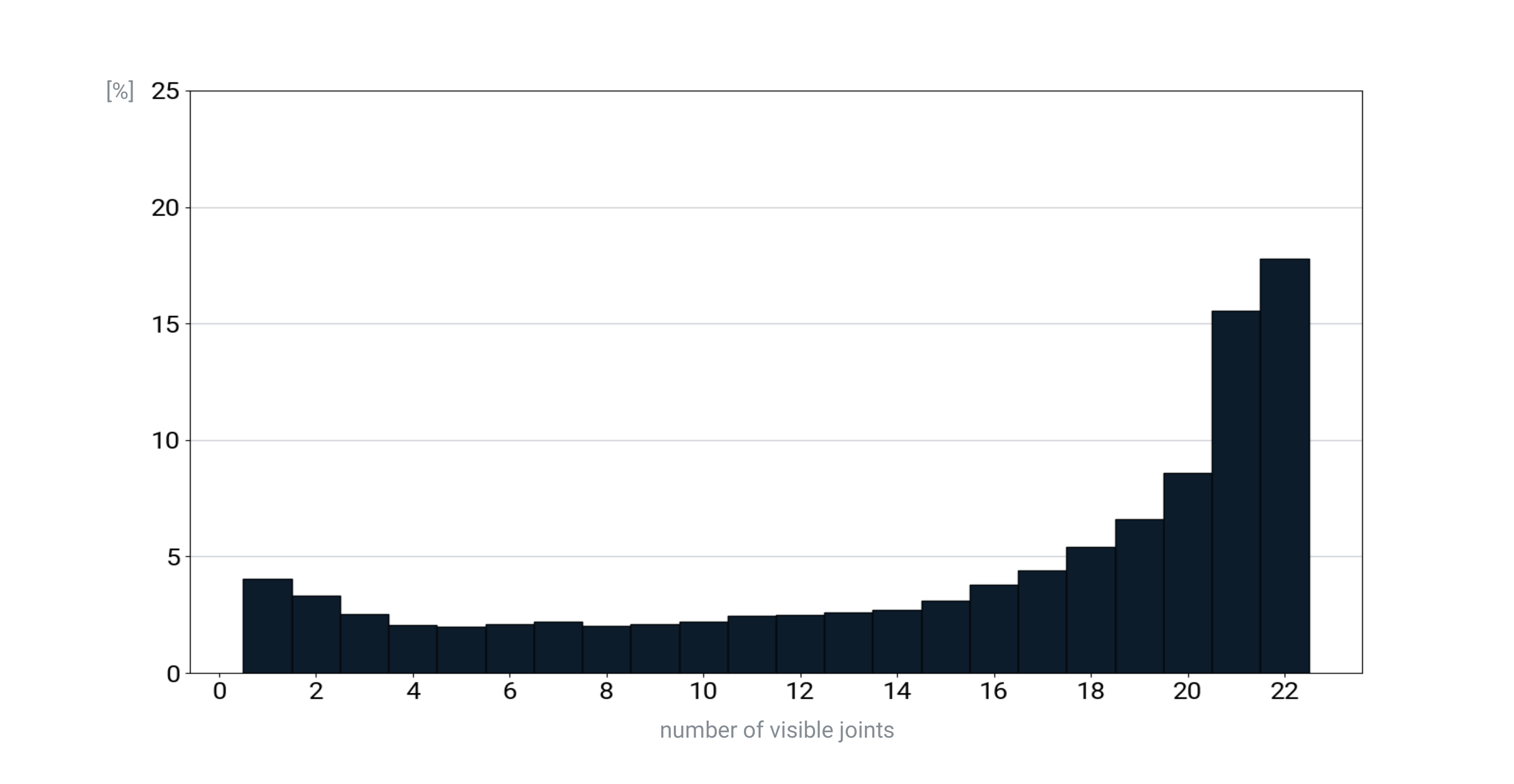}
         \caption{Visibility distribution of every annotated pedestrian reported as the number of visible joints.}
         \label{fig:visible}
     \end{subfigure}
\\[1px]
\caption{Additional statistics of the \motsynth dataset. Each distribution is calculated considering all pedestrians with at least one visible joint.}
\label{fig:histo}
\vspace{0.1cm}
\end{center}
\end{figure*}

We present a more detailed analysis of \motsynth in Fig~\ref{fig:histo}. In Fig.~\ref{fig:height} we plot the distribution of the bounding box heights expressed in pixels. As can be seen, $50\%$ of the bounding boxes are between $0$ and $95$ pixels. Only $2\%$ of them are higher than $613$ pixel. This clearly shows that \motsynth has been designed specifically for surveillance applications.

In Fig.~\ref{fig:bbox} we show the distribution of the number of bounding boxes per frame, ranging between $0$ and $125$ with a mean of $29.50$ and a standard deviation of $17.12$. The distribution is well balanced as peak values hardly reach a frequency of $2.5\%$.

In Fig.~\ref{fig:distance}, we plot the distance distribution of each pedestrian computed as the distance between the camera and the head joint expressed in meters. The average camera distance is $28.49$ meters, while the standard deviation is $20.33$ meters. Half of the annotations appear in $23$m range from the camera. Again, the peaks of the distribution never exceed $3\%$ showing good data balance.

In Fig.~\ref{fig:visible}, we plot pedestrian visibility distribution. It is calculated by counting the number of not occluded body joints, i.e., joints that are not obstructed by objects or other pedestrians and that are thus completely visible. \motsynth provide the annotation for 22 body joint, thus, a person is completely visible only if all his 22 joints are not occluded. The plot clearly shows that \motsynth is highly crowded as the percentage of completely visible pedestrians is less than $20\%$.

\section{Data Volume and Diversity}
\label{sec:volume-diversity}

In the main paper, we discussed the impact of the data sampling rate based on the Faster R-CNN detector~\cite{Ren15NIPS}. Here, we provide this analysis for all object detectors we experiment with in Tab.~\ref{tab:frcnn_fps}. 

YOLOv3 requires $104$k images to perform favourably \wrt COCO. Moreover, higher sampling rate is always beneficial both in term of AP and MODA. For CenterNet, the sampling rate does not impact the AP. For MODA, on the other hand, higher data volume seems to be beneficial. 

It is interesting to note that CenterNet is able to surpass COCO training with only $17$k images. Moreover, it is clear that visual diversity is crucial as split 4 with 1/60 sampling rate (17k images) surpasses the split 3 with 1/10 sampling rate (52k images). 

Results on Faster R-CNN are even more evident. With only $9$k images we obtain higher AP \wrt real data training. However, results seem to saturate with bigger splits. For both YOLOv3 and Faster R-CNN, split 2 with 1/10 sampling rate (24k frames) and split 3 and 4 with 1/60 sampling rate (9k and 17k frames respectively) obtain similar performance. 

This shows that volume and diversity are equally important. In general, visual diversity and data volume are equally important to achieve competitive results as the best performance is always obtained when diversity and volume are maximized.

\begin{table}[ht]
\center
\tabcolsep=0.11cm
\resizebox{\columnwidth}{!}{
\begin{tabular}{c|l| c c c c c c}
\toprule

& Method & MOTA $\uparrow$ & MOTP $\uparrow$ & IDF1 $\uparrow$ & FP $\downarrow$ & FN $\downarrow$ & IDS $\downarrow$ \\ [0.5ex] 

\midrule

\midrule
\parbox[t]{3mm}{\multirow{7}{*}{\rotatebox[origin=c]{90}{MOT20-04}}}
& Tracktor-\motsynth & 50.7 & 75.5 & 42.6 & 7383 & 125803 & 1963 \\
& CenterTrack-\motsynth & 41.7 & 74.5 & 38.7 & 15154 & 142557 & 2152 \\
& CenterTrack$^{\ddagger}$~\cite{zhou20ECCV} & 54.9 & \textbf{81.1} & 43.7 & \textbf{2187} & 118918 & 2641 \\
& Tracktorv2~\cite{Bergmann19ICCV} & 72.7 & 80.1 & 65.4 & 2855 & 71164 & 739 \\
& MPNTrack~\cite{braso2020learning} & \textbf{77.0} & 79.6 & 71.2 & 7459 & \textbf{55204} & \textbf{506} \\
& LPC~\cite{dai2021learning} & 75.7 & 80.6 & \textbf{75.7} & 4180 & 61864 & 648  \\
& SORT20~\cite{bewley2016simple} & 59.5 & 81.0 & 56.7 & 3206 & 106117 & 1643  \\

\midrule
\parbox[t]{3mm}{\multirow{7}{*}{\rotatebox[origin=c]{90}{MOT20-06}}}
& Tracktor-\motsynth & 35.5 & 73.9 & 33.7 & 4594 & 80171 & 871 \\
& CenterTrack-\motsynth & \textbf{40.8} & 71.8 & 35.3 & 12448 & \textbf{64330} & 1748 \\
& CenterTrack$^{\ddagger}$~\cite{zhou20ECCV} & 26.4 & 75.9 & 29.0 & 17481 & 78371 & 1881 \\
& Tracktorv2~\cite{Bergmann19ICCV} & 30.1 & \textbf{78.8} & 33.2 & \textbf{1745} & 90509 & 512 \\
& MPNTrack~\cite{braso2020learning} & 36.0 & 77.1 & 39.8 & 4831 & 79649 & \textbf{425} \\
& LPC~\cite{dai2021learning} & 35.3 & 77.7 & \textbf{43.2} & 3503 & 81891 & 499  \\
& SORT20~\cite{bewley2016simple} & 23.7 & 73.1 & 29.5 & 12309 & 87352 &  1640 \\

\midrule
\parbox[t]{3mm}{\multirow{7}{*}{\rotatebox[origin=c]{90}{MOT20-07}}}
& Tracktor-\motsynth & 52.5 & 77.5 & 50.9 & 509 & 15009 & 194 \\
& CenterTrack-\motsynth & 53.5 & 74.6 & 46.3 & 3082 & \textbf{11785} & 539 \\
& CenterTrack$^{\ddagger}$~\cite{zhou20ECCV} & 45.2 & 80.9 & 41.9 & 1101 & 16728 & 303 \\
& Tracktorv2~\cite{Bergmann19ICCV} & 50.1 & \textbf{81.1} & 49.6 & 252 & 16127 & 146 \\
& MPNTrack~\cite{braso2020learning} & \textbf{57.4} & 79.5 & \textbf{59.9} & 906 & 13061 & 120 \\

& LPC~\cite{dai2021learning} & 50.8 & 79.3 & 58.9 & \textbf{229} & 15921 &  \textbf{124} \\
& SORT20~\cite{bewley2016simple} & 48.5 & 77.6 & 47.3 & 1032 & 15666 &  360 \\
\midrule
\parbox[t]{3mm}{\multirow{7}{*}{\rotatebox[origin=c]{90}{MOT20-08}}}
& Tracktor-\motsynth & \textbf{29.4} & 73.2 & 33.5 & 3447 & 50831 & 439 \\
& CenterTrack-\motsynth & 24.6 & 68.7 & 31.4 & 16382 & \textbf{40602} & 1433 \\
& CenterTrack$^{\ddagger}$~\cite{zhou20ECCV} & 9.0 & 73.9 & 25.8 & 19736 & 49828 & 948 \\
& Tracktorv2~\cite{Bergmann19ICCV} & 21.0 & \textbf{78.8} & 27.2 & \textbf{2078} & 58880 & 251 \\
& MPNTrack~\cite{braso2020learning} & 25.9 & 77.3 & 36.1 & 3757 & 53470 & \textbf{159} \\
& LPC~\cite{dai2021learning} & 25.8 & 76.3 & \textbf{37.4} & 3814 & 53380 & 291  \\
& SORT20~\cite{bewley2016simple} & 13.1 & 70.9 & 24.2 & 10974 & 55559 & 827  \\

\bottomrule
\end{tabular}}
\vspace{-7pt}
\caption{Per-sequence benchmark results on MOT20.}
\label{tab:mot20_detailed}
\end{table}

\section{Implementation Details}
\label{sec:implementation}

\subsection{Object Detection Experiments}

\paragraph{YOLOv3.} For YOLOv3, we used Darknet backbone~\cite{redmon2016you}. 
We trained our model on \motsynth for $200,000$ iterations using the batch-size of $16$.  We resize input images to $608\times608$. We used the Ultralytics implementation~\cite{ultralytics} with default hyperparameters. For the evaluation, we used a confidence threshold of $0.4$ when testing on MOT17 and MOT20. 

\paragraph{CenterNet.} For CenterNet, we used DLA-34 backbone~\cite{yu2018deep} and used used the official implementation of CenterNet~\cite{cnet}. 
We trained on \motsynth for $100,000$ iterations using batch size $32$ (we used two GPUs). 
During the inference, we used a confidence threshold of $0.3$ when testing on MOT17 and a confidence threshold of $0.1$ when testing on MOT20. 

\paragraph{Faster R-CNN and Mask R-CNN.} For Faster R-CNN, we use a ResNet50~\cite{He16CVPR} backbone with FPN~\cite{Lin_2017_CVPR} (Detectron2~\cite{Detectron2018} implementation). We train models on \motsynth for $35,000$ iterations and use default Detectron2 hyperparameters. 
To avoid overfitting, we freeze all the backbone blocks except for the last one. 
For fine-tuning, we follow~\cite{Bergmann19ICCV} and train our models for 30 additional epochs on the respective dataset. Similarly, we follow the same training scheme and use the same hyperparameters for Mask R-CNN.

\subsection{Person Re-Identification Experiments}

For ReID, we follow~\cite{transfer_reid}: we freeze all CNN layers and pre-train the fully connected layers for 5 epochs. 
We then train our entire models for 55 additional epochs using Adam optimizer (citation needed) and a learning rate of $0.004$. We resize images to $128x56$ and use random cropping and flipping data augmentation techniques. 

\subsection{Multi-Object Tracking Experiments}

\begin{table}[h]
\center
\tabcolsep=0.11cm
\resizebox{\columnwidth}{!}{
\begin{tabular}{c|l| c c | c c c}
\toprule
& Method & MOTA $\uparrow$ & IDF1 $\uparrow$ & FP $\downarrow$ & FN $\downarrow$ & IDS $\downarrow$  \\ [0.5ex] 
\midrule
\parbox[t]{3mm}{\multirow{10}{*}{\rotatebox[origin=c]{90}{Public}}} 
& Tracktor-\motsynth & 56.9 & 56.9 & 20852 & 220273 & 2012 \\
& Tracktor-\motsynth + FT & 59.1 & 58.8 & 22231 & 206062 & 2323 \\
& Tracktor~\cite{Bergmann19ICCV} & 53.5 & 52.3 & 12201 & 248047 & 2072  \\
& Tracktorv2~\cite{Bergmann19ICCV} & 56.3 & 55.1 & \textbf{8866} & 235449 & 1987 \\
\cline{2-7}
& CenterTrack-\motsynth & 59.7 & 52.0 & 39707 & 181471 & 6035 \\
& CenterTrack-\motsynth + FT & \textbf{65.1} & 57.9 & 11521 & \textbf{180901} & 4377 \\
& CenterTrack~\cite{zhou20ECCV} & 61.5 & 59.6 & 14076 & 200672 & 2583 \\
\cline{2-7}

& Lif\_T~\cite{hornakova20ICML} & 60.5 & 65.6 & 14966 & 206619 & 1189  \\
& LPC~\cite{dai2021learning} & 59.0 &\textbf{ 66.8} & 23102 & 206948 & \textbf{1122}  \\
& MPNTrack~\cite{braso2020learning} & 58.8 & 61.7 & 17413 & 213594 & 1185  \\

\midrule
\midrule
\parbox[t]{3mm}{\multirow{3}{*}{\rotatebox[origin=c]{90}{Private}}} 
& CorrTracker~\cite{Wang_2021_CVPR} & \textbf{76.5} & \textbf{73.6} & 29808 & \textbf{99510} & 3369 \\
& FairMOTv2~\cite{zhang2020simple} & 73.7 & 72.3 & 27507 & 117477 & \textbf{3303} \\
& TraDeS~\cite{Wu_2021_CVPR} & 69.1 & 63.9 & \textbf{20892} & 150060 & 3555 \\
\bottomrule
\end{tabular}}
\vspace{-7pt}
\caption{Detailed Benchmark results on MOT17.}
\label{tab:mot-bench}
\end{table}

\paragraph{MOT.} For \textit{CenterTrack}~\cite{zhou20ECCV}, we follow the training schemes explained in Section 4.4 of the main paper. 
We fine-tune our network for 30 epochs for MOT17 and 70 epochs for the MOT20 dataset for the fine-tuning experiments. 
We train and evaluate our models using the same hyperparameters as reported by~\cite{zhou20ECCV}. 
For \textit{Tracktor}~\cite{Bergmann19ICCV}, we follow the setting described for Faster R-CNN and ReID, as no additional training is required: \textit{Tracktor} leverages bounding box regression head of Faster R-CNN detector, trained on static images.

\paragraph{MOTS.} We adapt our Mask R-CNN model, trained on \motsynth, by using bounding box regression mechanism for tracking and mask segmentation head provides segmentation masks (Mask R-CNN Tracktor ($\dagger$)). For all experiments and the benchmark submission, we use the same ReID network and hyperparameters as reported in~\cite{Bergmann19ICCV}. 

\section{Detailed Benchmark Results}
\label{sec:mot20_detailed}

In Tab.~\ref{tab:mot20_detailed} we present the detailed MOT20 benchmark results for each sequence and analyze how Tracktor and CenterTrack (trained only on \motsynth) compare with the state-of-the-art trackers in extremely crowded scenes. In addition to published models, we train and evaluate CenterTrack on MOT20 (denoted with $\ddagger$), following the training procedure of \cite{zhou20ECCV}.We are interested in comparing existing models trained on different datasets. Therefore, we use the default CenterTrack hyperparameters. %

We observe that in sequence MOT20-04, Tracktor-\motsynth and CenterTrack-\motsynth are not on-par with Tracktorv2, MPNTrack and LPC. This is likely because the sequences with near-bird's-eye viewpoints (similar to MOT20-04) are rare in \motsynth dataset. However, in all other MOT20 sequences, Tracktor and CenterTrack only trained on synthetic data outperform Tracktorv2 with a significant margin and are on-par with the state-of-the-art. Fine-tuning these models on MOT20 further improves their performance, as reported in Section 4.6 of the paper. These experiments indicate that top-performing tracking models can be trained on synthetic data even in extremely dense scenarios. 

\begin{table}[ht]
\center
\tabcolsep=0.11cm
\resizebox{\columnwidth}{!}{
\begin{tabular}{c|l| c c | c c c}
\toprule
& Method & MOTA $\uparrow$ & IDF1 $\uparrow$ & FP $\downarrow$ & FN $\downarrow$ & IDS $\downarrow$  \\ [0.5ex] 
\midrule
\parbox[t]{3mm}{\multirow{8}{*}{\rotatebox[origin=c]{90}{Public}}} 
& Tracktor-\motsynth & 43.7 & 39.7 & 15933 & 271814 & 3467 \\
& Tracktor-\motsynth + FT & 56.5 & 52.8 & 11377 & 211772 & 1995 \\
& Tracktorv2~\cite{Bergmann19ICCV} & 52.6 & 52.7 & \textbf{6930} & 236680 & 1648 \\
\cline{2-7}
& CenterTrack-\motsynth & 39.7 & 37.2 & 47066 & 259274 & 5872 \\
& CenterTrack-\motsynth + FT & 41.9 & 38.2 & 36594 & 258874 & 5313 \\
\cline{2-7}
& MPNTrack~\cite{braso2020learning} & \textbf{57.6 } & 59.1 & 16953 & \textbf{201384} & \textbf{1210}  \\
& LPC~\cite{dai2021learning} & 56.3 & \textbf{62.5} & 11726 & 213056 & 1562  \\
& SORT20~\cite{bewley2016simple} & 42.7 & 45.1 & 27521 & 264694 & 4470  \\
\midrule
\midrule
\parbox[t]{3mm}{\multirow{3}{*}{\rotatebox[origin=c]{90}{Private}}} 
& JDMOTGNN~\cite{wang2020joint} & \textbf{67.1} & 67.5 & \textbf{31913} & 135409 & \textbf{3131} \\
& CorrTracker~\cite{Wang_2021_CVPR} & 65.2 & \textbf{69.1} & 79429 & 95855 & 5183 \\
& FairMOTv2~\cite{zhang2020simple} & 61.8 & 67.3 & 103440 & \textbf{88901} & 5243 \\
\bottomrule
\end{tabular}}
\vspace{-7pt}
\caption{Detailed Benchmark results on MOT20.}
\label{tab:mot-bench}
\end{table}

\end{document}